\documentclass{article}

\PassOptionsToPackage{numbers, compress}{natbib}
\usepackage[preprint]{neurips_2021}  %

\usepackage[utf8]{inputenc} 
\usepackage[T1]{fontenc}    
\usepackage{hyperref}       
\usepackage{url}            
\usepackage{booktabs}       
\usepackage{amsfonts}       
\usepackage{nicefrac}       
\usepackage{microtype}      
\usepackage{xcolor}         
\usepackage{multirow}

\usepackage{microtype}
\usepackage{graphicx}
\usepackage{booktabs} 
\usepackage{caption}
\usepackage{subcaption}

\graphicspath{figures/}

\usepackage{hyperref}
\usepackage{amsmath}
\usepackage{adjustbox}

\usepackage{xcolor}
\usepackage{xspace}
\newtheorem{theorem}{Theorem}

\usepackage{amsmath,amsfonts,bm}
\usepackage{dsfont}
\usepackage{centernot}

\newcommand{\mb}[1]{\mathbf{#1}}

\newcommand{\indicator}{\mathds{1}}

\def\eqref#1{equation~\ref{#1}}

\def\1{\bm{1}}

\def\vf{{\bm{f}}}

\def\vw{{\bm{w}}}
\def\vx{{\bm{x}}}
\def\vy{{\bm{y}}}

\def\mJ{{\bm{J}}}

\def\mX{{\bm{X}}}

\DeclareMathAlphabet{\mathsfit}{\encodingdefault}{\sfdefault}{m}{sl}
\SetMathAlphabet{\mathsfit}{bold}{\encodingdefault}{\sfdefault}{bx}{n}

\title{On Stein Variational Neural Network Ensembles}

\author{Francesco D'Angelo \\
  ETH Zürich\\
  Zürich, Switzerland \\
  \texttt{fdangelo@ethz.ch} \\
  \And
  Vincent Fortuin \\
  ETH Zürich\\
  Zürich, Switzerland \\
  \texttt{fortuin@inf.ethz.ch} \\
  \And
  Florian Wenzel \\
  Humboldt University of Berlin\\
  Berlin, Germany\\
  \texttt{wenzelfl@hu-berlin.de}\\
}

\begin{document}

\maketitle

\begin{abstract}
Ensembles of deep neural networks have achieved great success recently, but they do not offer a proper Bayesian justification.
Moreover, while they allow for averaging of predictions over several hypotheses, they do not provide any guarantees for their diversity, leading to redundant solutions in function space.
In contrast, particle-based inference methods, such as Stein variational gradient descent (SVGD), offer a Bayesian framework, but rely on the choice of a kernel to measure the similarity between ensemble members.
In this work, we study different SVGD methods operating in the weight space, function space, and in a hybrid setting. 
We compare the SVGD approaches to other ensembling-based methods in terms of their theoretical properties and assess their empirical performance on synthetic and real-world tasks.
We find that SVGD using functional and hybrid kernels can overcome the limitations of deep ensembles. It improves on functional diversity and uncertainty estimation and approaches the true Bayesian posterior more closely.
Moreover, we show that using stochastic SVGD updates, as opposed to the standard deterministic ones, can further improve the performance.
\end{abstract}

\section{Introduction}
\label{sec:intro}

Ensembling methods for neural networks have achieved great success recently, both in terms of predictive performance \citep{lakshminarayanan2017simple, wenzel2020hyper} as well as uncertainty estimation \citep{ovadia2019can}.
It has even been argued that these methods can be viewed as an approximate way of performing Bayesian inference in neural networks \citep{wilson2020bayesian}.
However, while they do allow for the marginalization of predictions over several hypotheses, they do not specifically take a prior distribution into account, nor do they encourage diversity between hypotheses, in contrast to actual Bayesian neural networks (BNNs) \citep{mackay1992practical}.

    

One way to bridge the gap between deep ensembles and BNNs is through an inference method called Stein variational gradient descent (SVGD) \citep{liu2016stein}.
SVGD uses particles to approximate the Bayes posterior and therefore in practice also trains an ensemble of neural network models.
However, as opposed to standard deep ensembles, it uses an objective function that encourages diversity in the ensemble through the use of a kernel.
Moreover, it guarantees asymptotic convergence to the true posterior \citep{liu2017stein, korba2020non}.

While SVGD can na\"ively be defined using a kernel in the weight space of the neural network, this can lead to suboptimal results in the case of BNN inference, since the mapping from weights to functions is quite complex.
For instance, several neural network particles could have very different weights, but implement the exact same function, thus giving a false sense of diversity in the ensemble \citep{badrinarayanan2015understanding, fort2019deep}.
Moreover, while the conventional SVGD update is a deterministic function, it has recently been proposed that stochatic updates can improve the diversity even further \citep{gallego2018stochastic}.

\begin{figure*}
\centering
    \begin{subfigure}[b]{0.15\textwidth}
    \includegraphics[width=\linewidth,trim={1cm 1cm 1cm 0cm},clip]{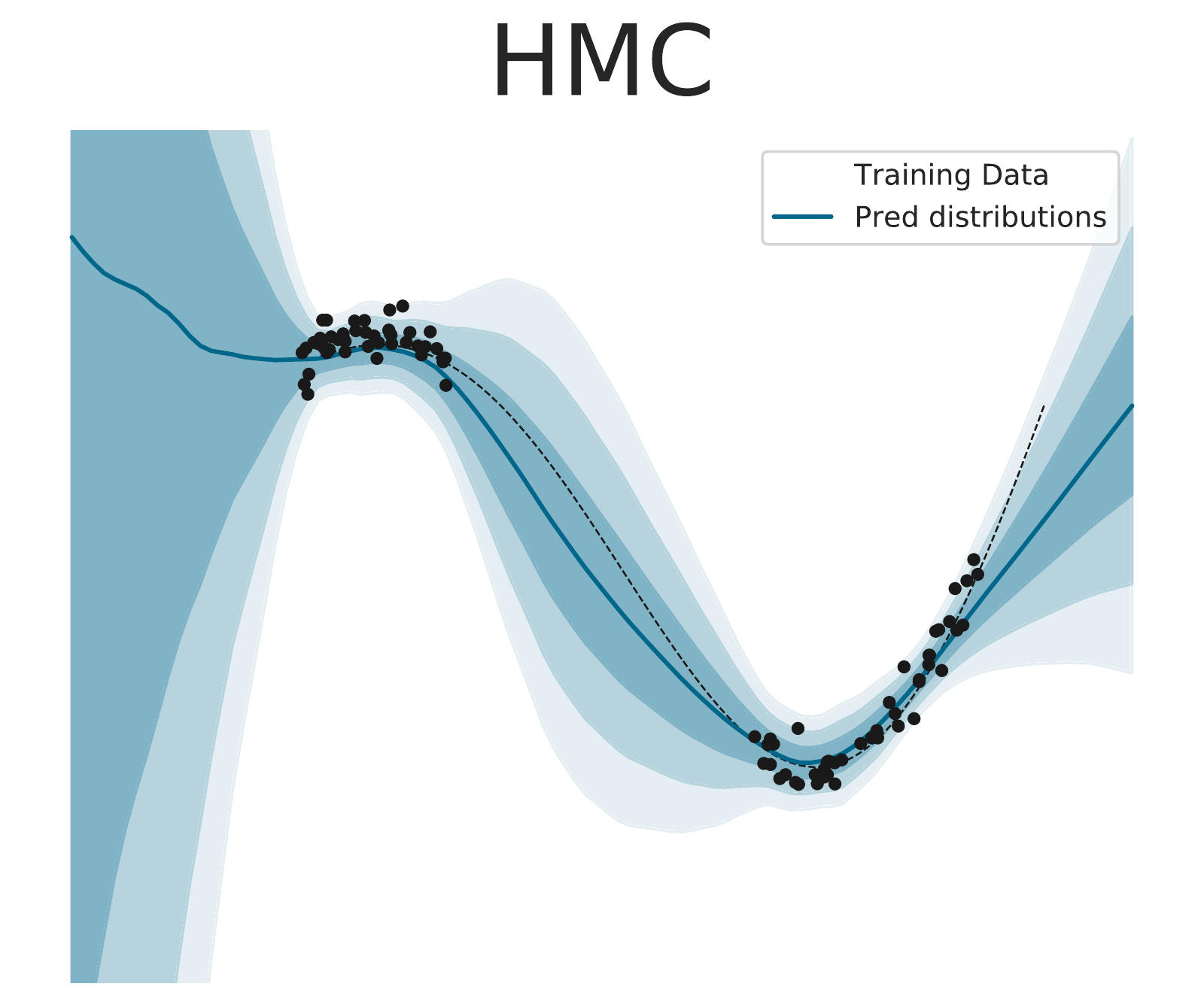}
    \end{subfigure}
    \begin{subfigure}[b]{0.15\textwidth}
    \includegraphics[width=\linewidth,trim={1cm 1cm 1cm 0cm},clip]{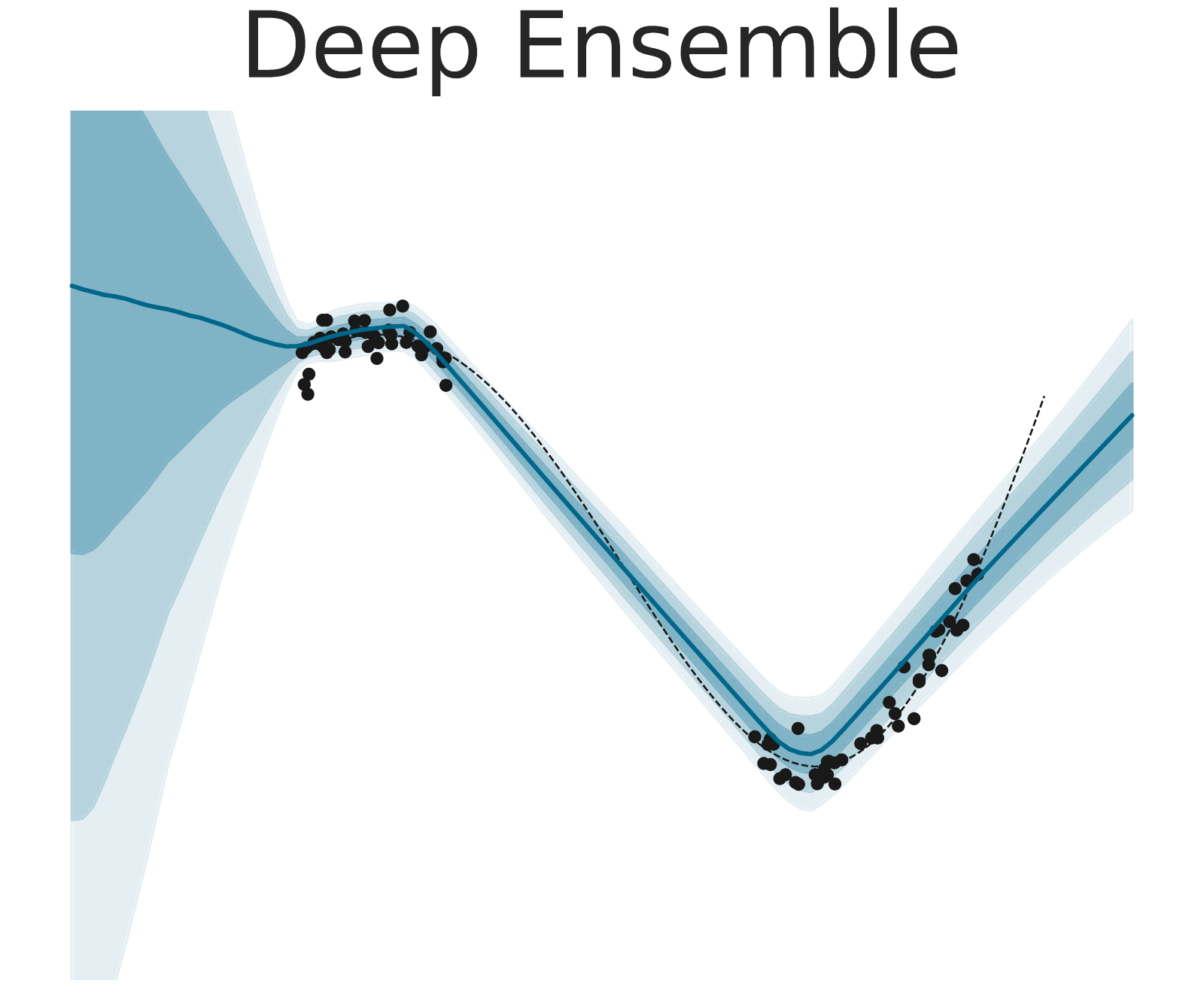}
    \end{subfigure}
    \begin{subfigure}[b]{0.15\textwidth}
    \includegraphics[width=\linewidth,trim={1cm 1cm 1cm 0cm},clip]{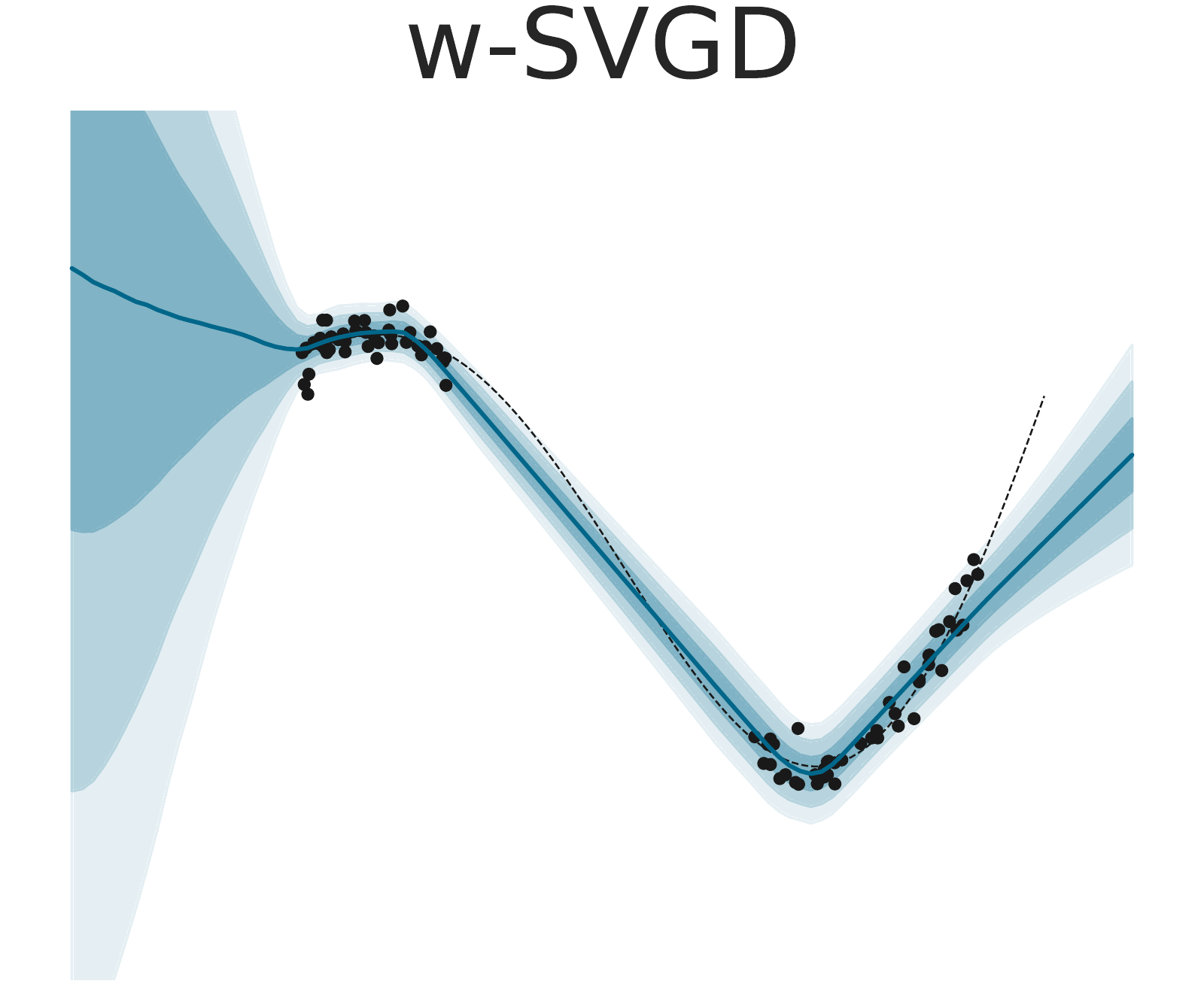}
    \end{subfigure}
    \begin{subfigure}[b]{0.15\textwidth}
    \includegraphics[width=\linewidth,trim={1cm 1cm 1cm 0cm},clip]{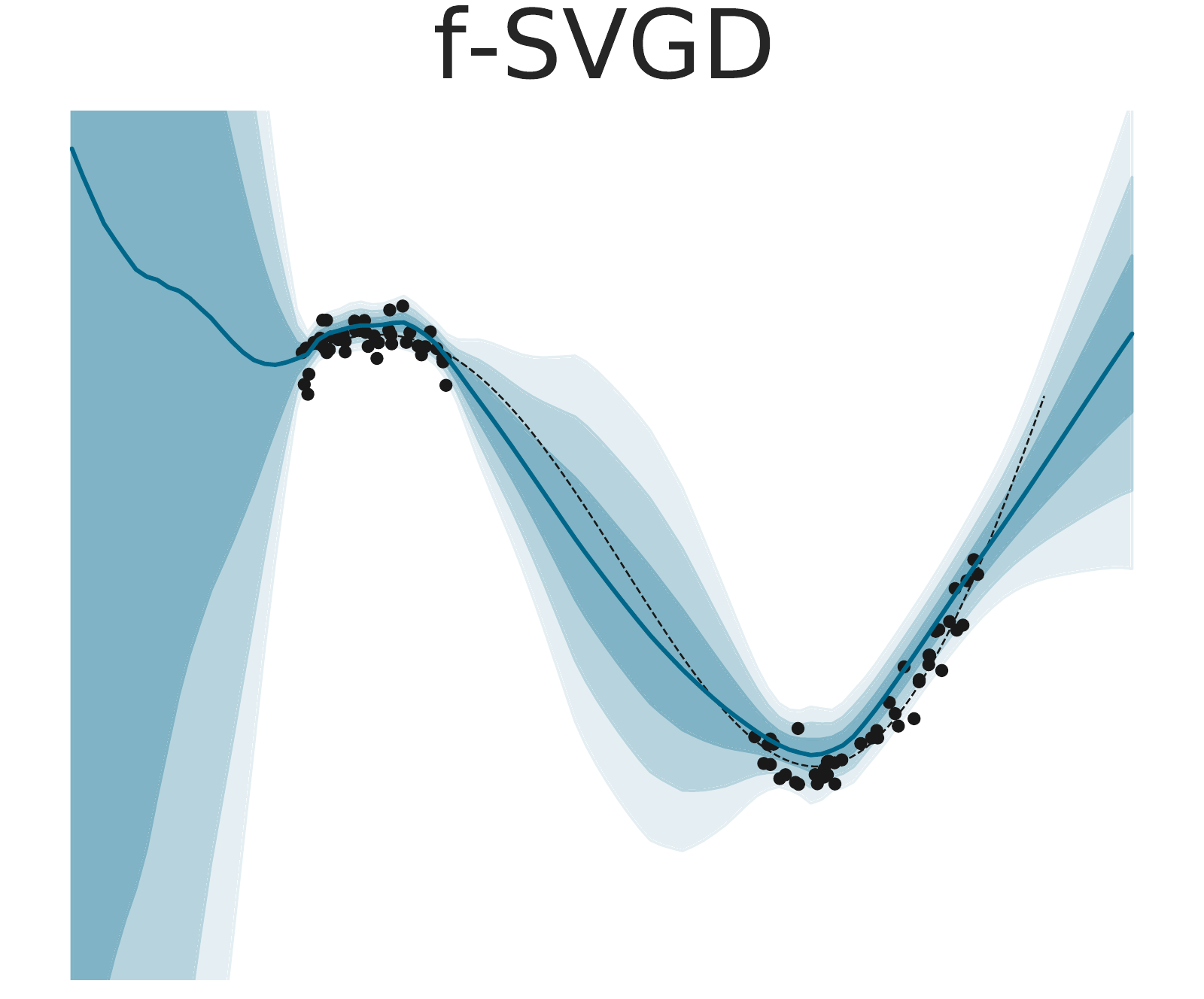}
    \end{subfigure}
    \begin{subfigure}[b]{0.15\textwidth}
    \includegraphics[width=\linewidth,trim={1cm 1cm 1cm 0cm},clip]{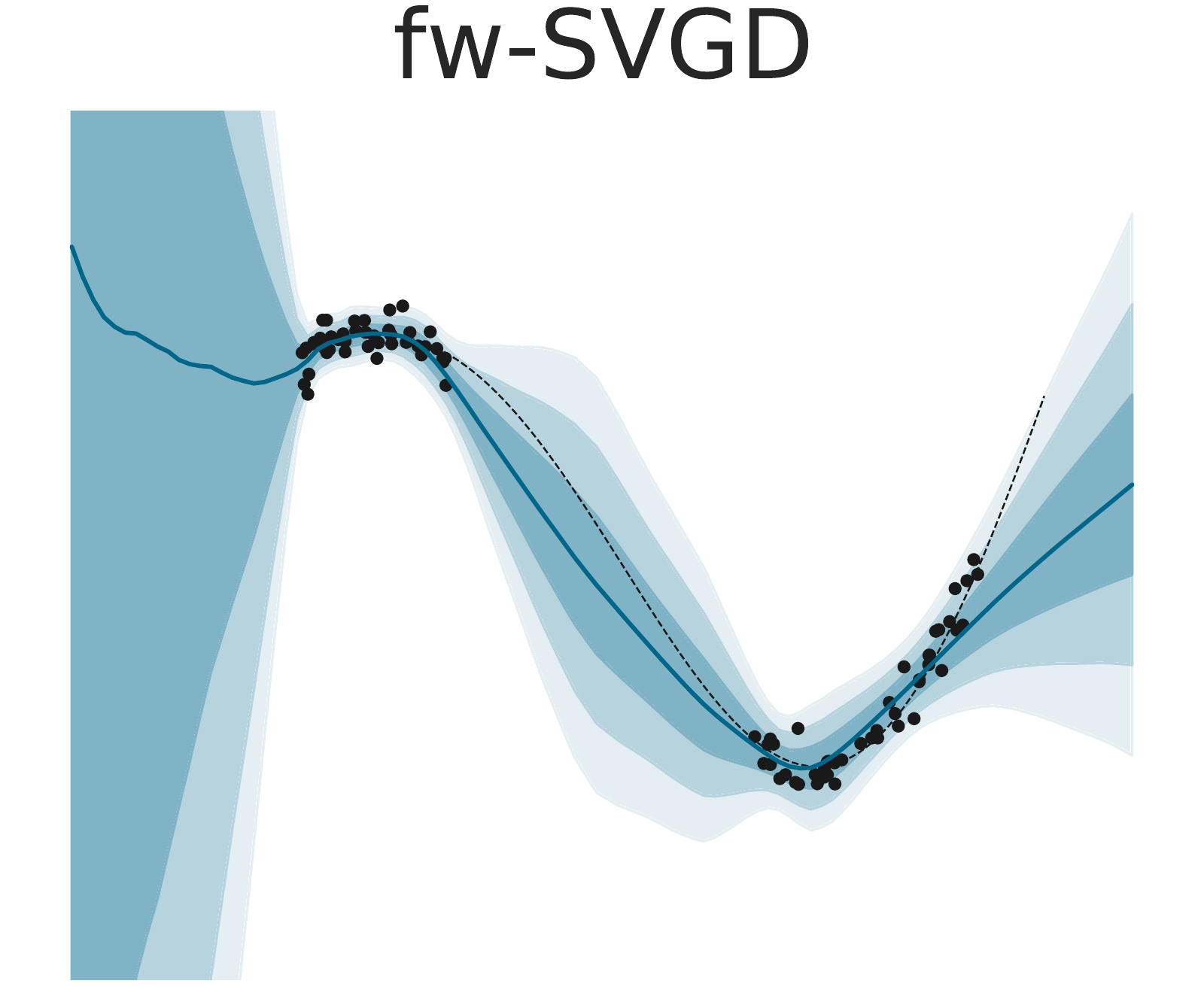}
    \end{subfigure}
    \begin{subfigure}[b]{0.15\textwidth}
    \includegraphics[width=\linewidth,trim={1cm 1cm 1cm 0cm},clip]{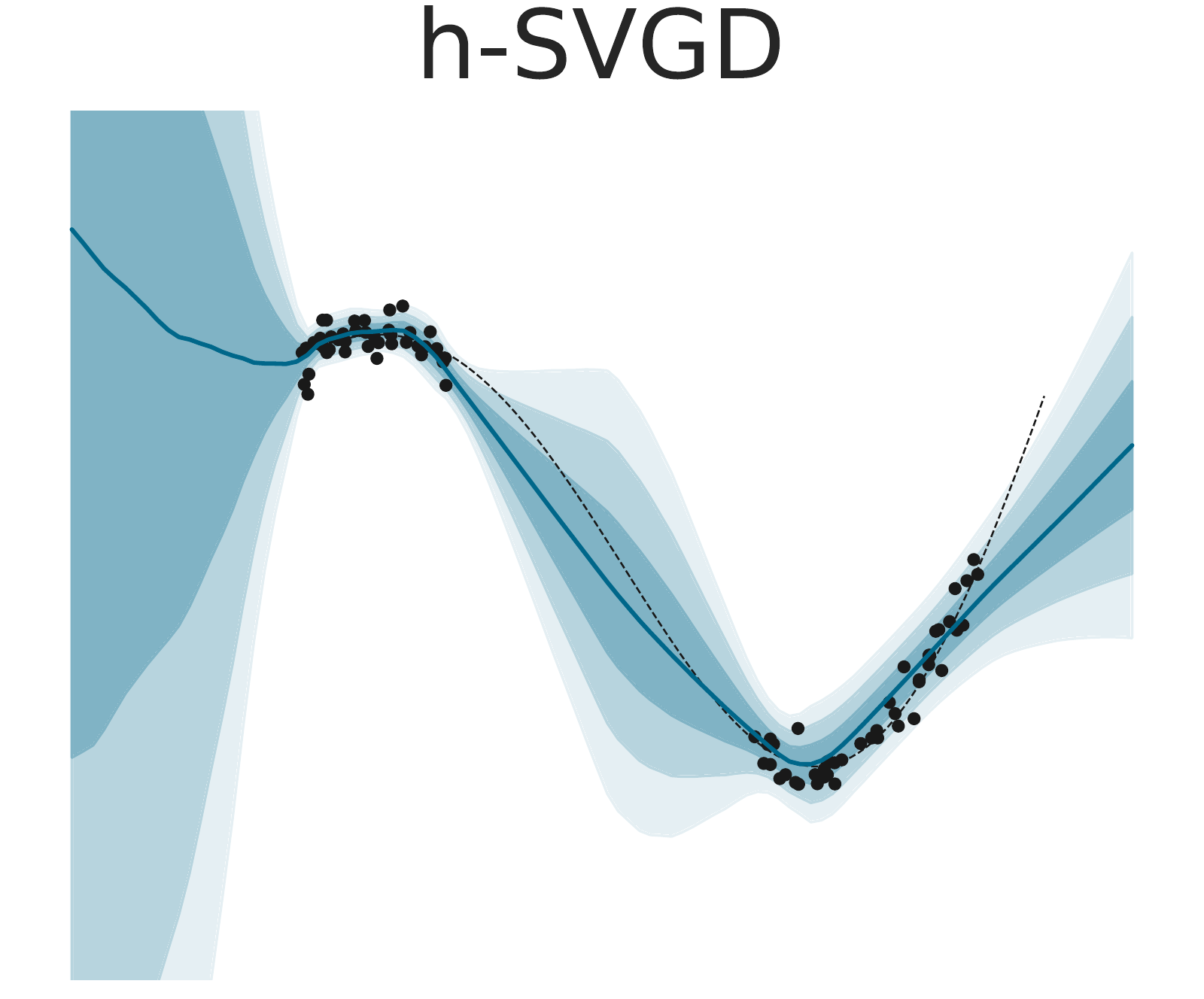}
    \end{subfigure}
    
    \hspace{3.8mm}
    \begin{subfigure}[b]{0.15\textwidth}
    \includegraphics[width=\linewidth,trim={3cm 3cm 3cm 3cm},clip]{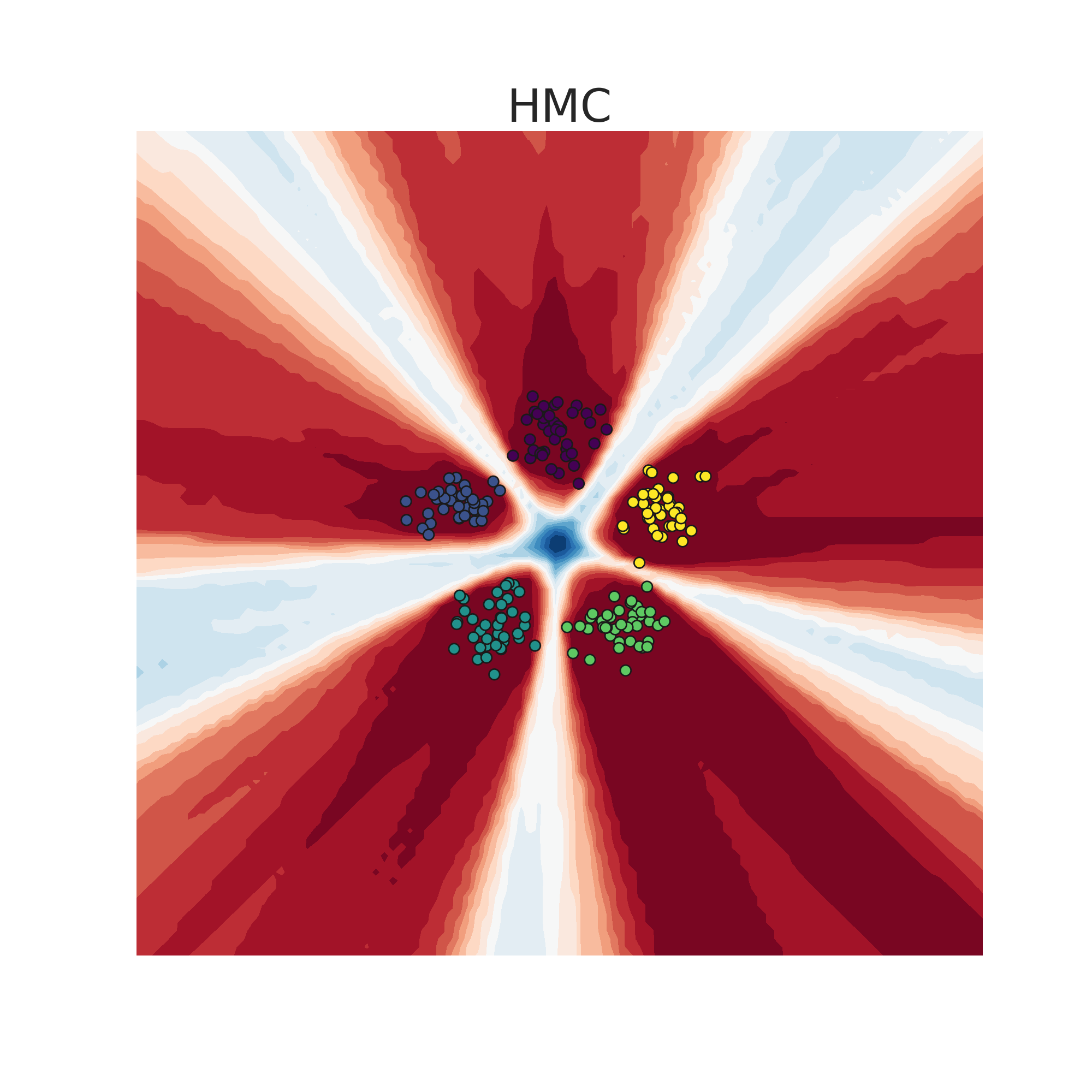}
    \end{subfigure}
    \begin{subfigure}[b]{0.15\textwidth}
    \includegraphics[width=\linewidth,trim={3cm 3cm 3cm 3cm},clip]{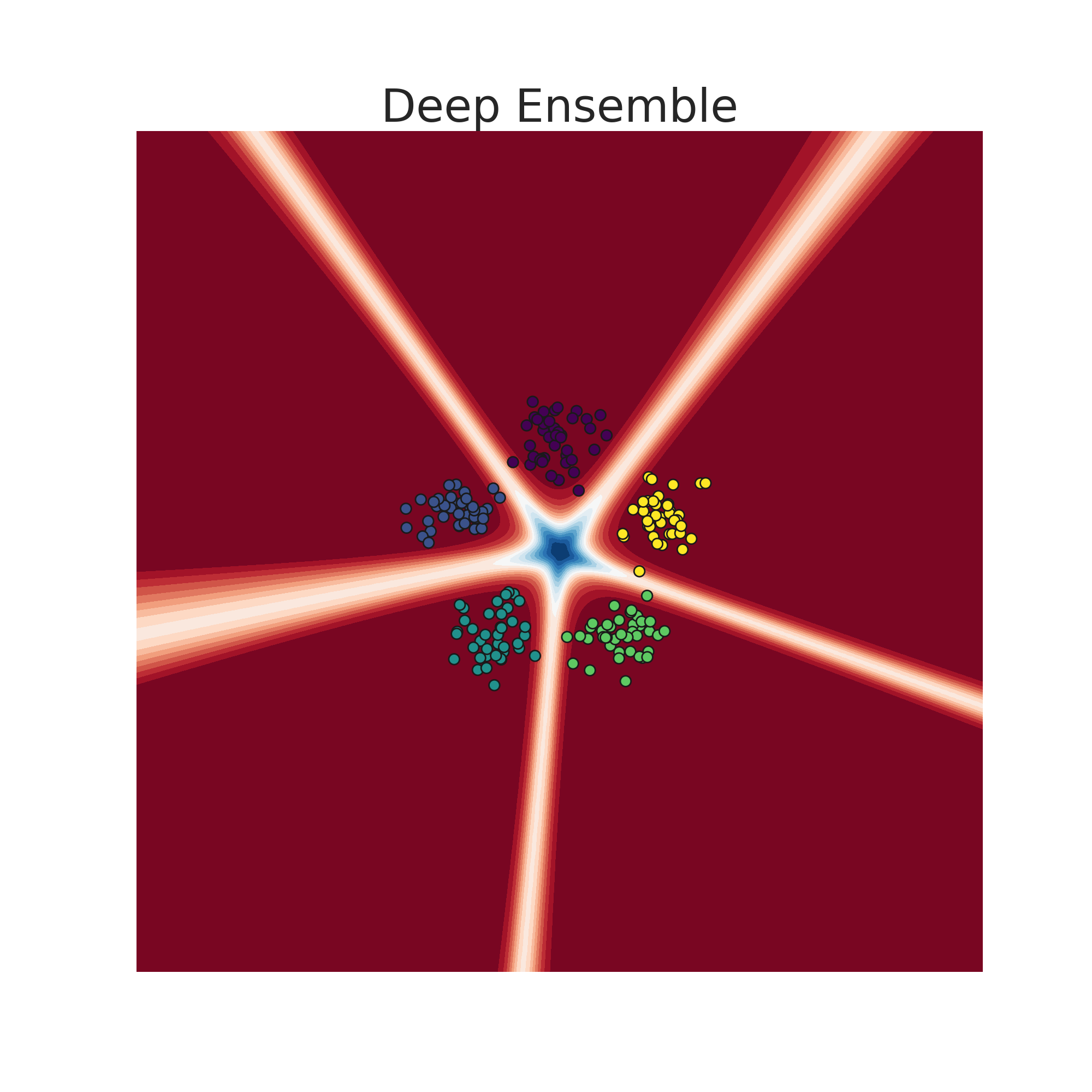}
    \end{subfigure}
    \begin{subfigure}[b]{0.15\textwidth}
    \includegraphics[width=\linewidth, trim={3cm 3cm 3cm 3cm},clip]{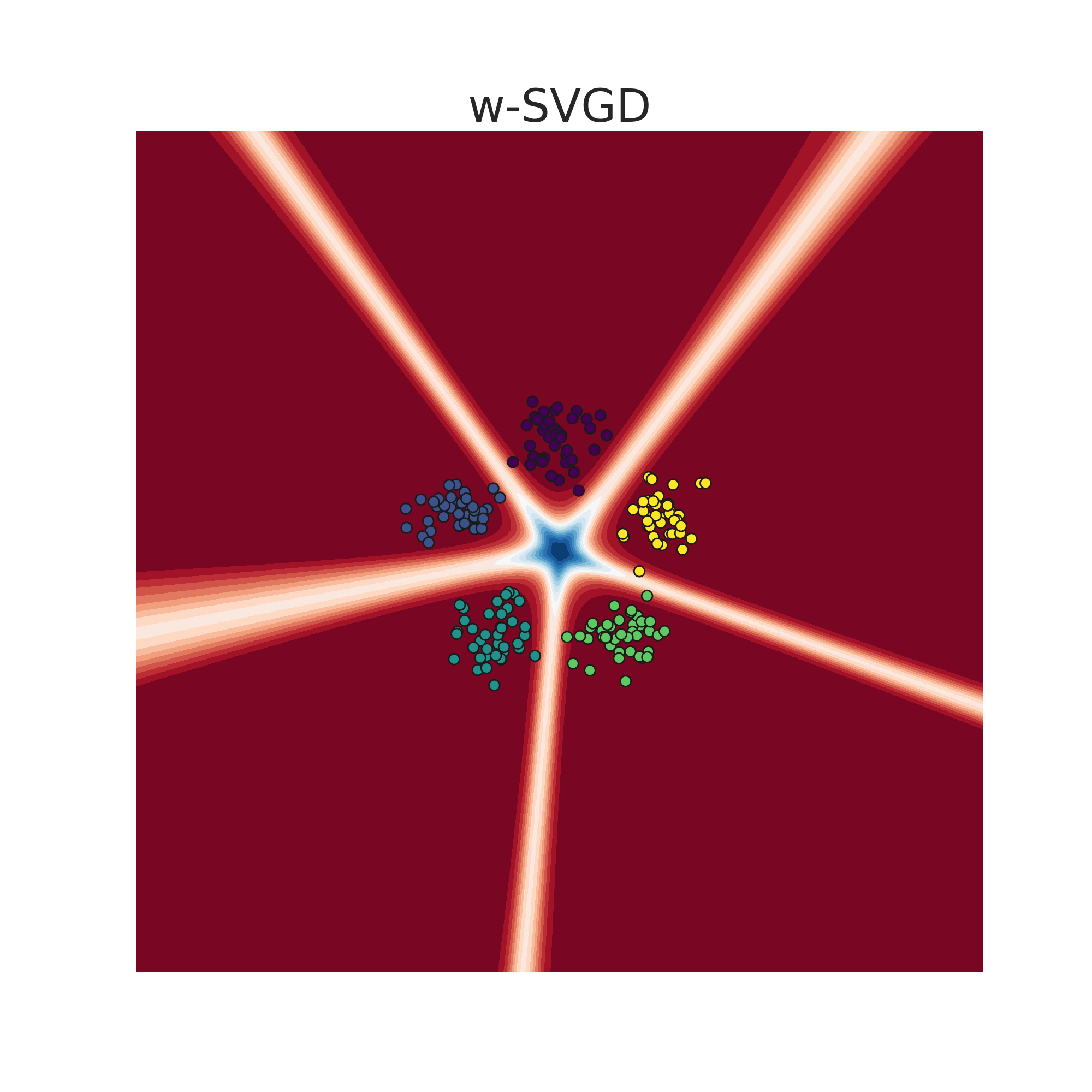}
    \end{subfigure}
    \begin{subfigure}[b]{0.15\textwidth}
    \includegraphics[width=\linewidth, trim={3cm 3cm 3cm 3cm},clip]{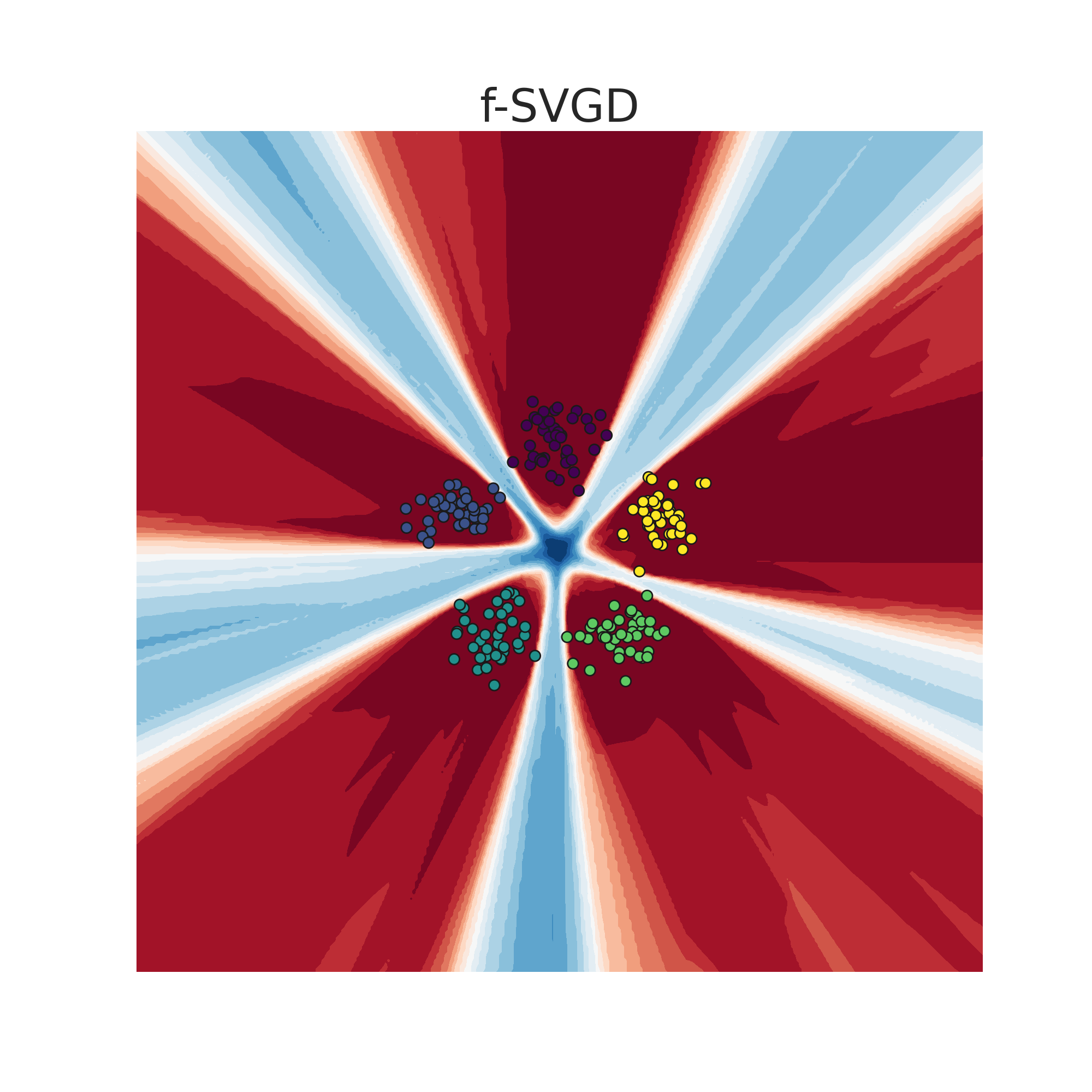}
    \end{subfigure}
    \begin{subfigure}[b]{0.15\textwidth}
    \includegraphics[width=\linewidth, trim={3cm 3cm 3cm 3cm},clip]{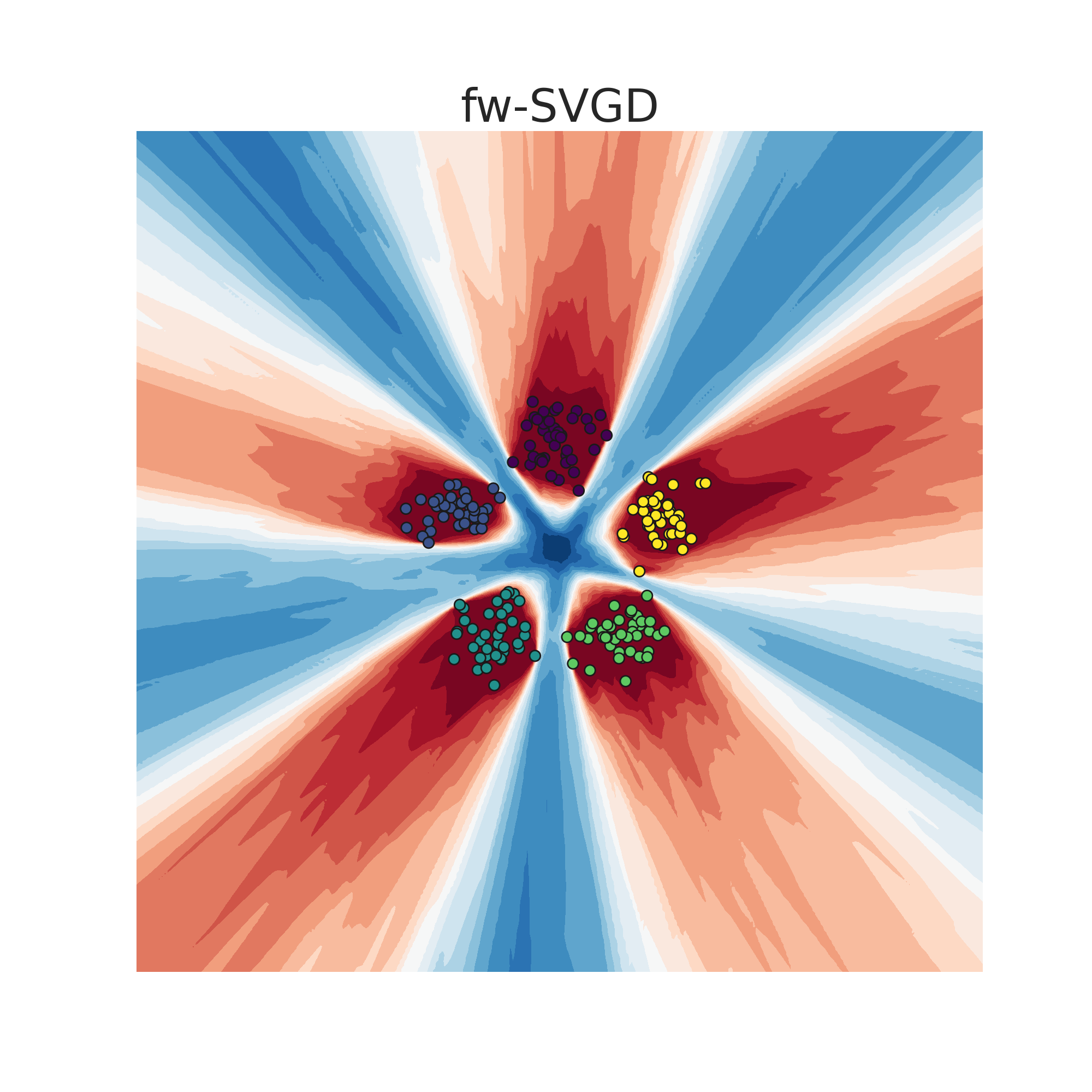}
    \end{subfigure}
    \begin{subfigure}[b]{0.15\textwidth}
    \includegraphics[width=\linewidth, trim={3cm 3cm 3cm 3cm},clip]{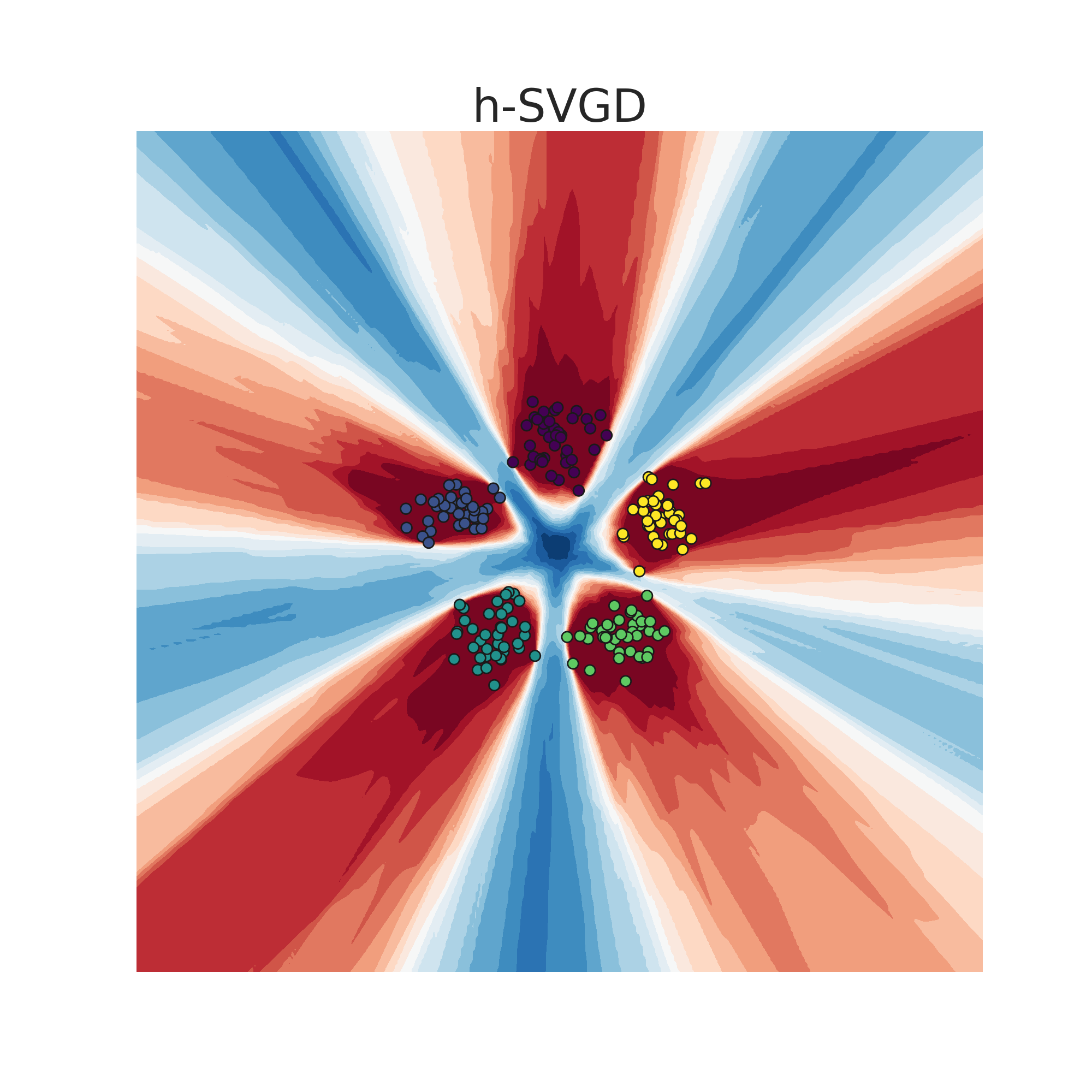}
    \end{subfigure}
    \begin{subfigure}[b]{0.027\textwidth}
    \includegraphics[width=\linewidth,trim={0.4cm 0.5cm 0cm 0.2cm},clip]{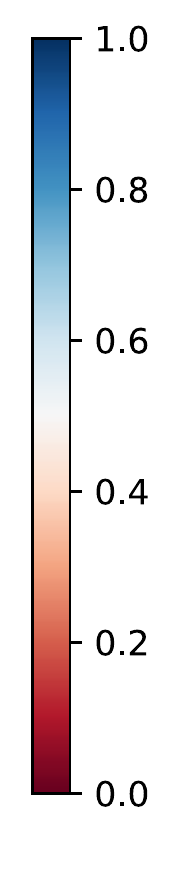}
    \vspace{-5mm}
    \end{subfigure}
\caption{
We compare different training methods for BNNs on synthetic data (details in section~\ref{sec:experiments}). In the regression setting (top) the shaded areas represent the standard deviation and in the classification setting (bottom) the entropy of the predictive posteriors. Deep ensembles and SVGD in weight space (w-SVGD) do not encourage diversity in the function space and fail to represent in-distribution uncertainty in regression and uncertainty away from the training data in classification.  In this work, we study variants of SVGD using weight-space, function-space, and hybrid kernels. The functional methods (f-SVGD and our new methods fw-SVGD and h-SVGD) improve diversity and uncertainty representation and closely approach the ground truth posterior approximated by HMC.}
\label{fig:toy_tasks}
\end{figure*}

In this work, we explore different ensembling and BNN sampling methods and assess their diversity between predictions. We compare SVGD methods on a spectrum between pure weight space and pure function space inference, using deterministic and stochastic updates, and characterize them theoretically and empirically. We also include two new approaches in our comparison and show that our hybrid h-SVGD method, that acts both in the weight and function space, and our fw-SVGD method, that fixes an issue with an existing functional SVGD approach \citep{wang2019function}, lead to more diverse ensembles and improved uncertainty estimation and out-of-distribution detection, as well as approaching the gold-standard Hamiltonian Monte Carlo posterior more closely.

We make the following contributions:

\begin{itemize}
    \item We describe a generalized framework for Stein variational neural network ensembles, which includes several existing methods (such as standard deep ensembles) as special cases.
    \item We characterize different kernel choices as well as deterministic and stochastic update rules within this framework in terms of their theoretical properties.
    \item We empirically compare the different methods on several synthetic and real-world data sets and highlight their strengths and weaknesses for various use cases.
    \item We find that SVGD variants with function-space kernels, including our newly proposed ones, can empirically lead to improved uncertainty estimation and out-of-distribution detection as well as approach the gold-standard posterior more closely.
    \item We also find that stochastic updates can improve performance over deterministic ones.
\end{itemize}

\section{Background}
\label{sec:background}

\subsection{Deep Neural Network Ensembles}
\label{sec:deep_ensembles}

Ensembles of neural networks have a long history \citep[e.g.,][]{lts-salglnn-90,hansen90, breiman1996bagging} and were recently revisited by \citet{lakshminarayanan2017simple} and coined \emph{deep ensembles}.
A deep ensemble is constructed by training several networks \emph{independently} and then combining their predictions.
While this method in principle allows to account for different hypotheses, it has been shown that it can still severely overfit \citep{rahaman2020uncertainty} and requires the injection of additional randomness to yield uncertainty estimates that are conservative with respect to the true Bayesian ones \citep{ciosek2019conservative}.
Since the quality of an ensemble hinges on the diversity of its members, many methods were proposed to improve the diversity \citep{wenzel2020hyper, snap17}.
However, in all these methods, the members do not interact with each other and can thus end up converging to the same optimum in function space \citep{fort2019deep}.
In contrast, our work directly focuses on improving the diversity of ensembles (i.e., the complementarity of predictions) by combining them with ideas from SVGD.

\subsection{Bayesian Neural Networks}
\label{sec:bnns}

In supervised deep learning, we typically consider  a likelihood function $p(\vy | f(\vx; \vw))$ (e.g., Gaussian for regression or Categorical for classification) parameterized by a neural network $f(\vx; \vw)$ and training data $\mathcal{D} = \{(x_i,y_i)\}_{i=1}^n$. 
In Bayesian neural networks (BNNs), we are interested in the  posterior distribution of all likely networks given by
$
    p(\vw | \vx, \vy) \propto p(\vy | f(\vx; \vw)) \, p(\vw)
$,
where $p(\vw)$ is the prior distribution over weights, which is often chosen to be standard Gaussian.
Crucially, when making a prediction on a test point $\vx^*$, in the Bayesian approach we do not only use a single parameter $\hat{\vw}$ to predict $\vy^* = f(\vx^*; \hat{\vw})$, but we marginalize over the whole posterior, thus taking all possible explanations of the data into account:
\begin{equation}
\label{eqn:bayes_predictive}
    p(\vy^* | \vx^*, \vx, \vy) = \int p(\vy^* | f(\vx^*; \vw)) \, p(\vw | \vx, \vy) \, \mathrm{d}\vw
\end{equation}


The main challenge for inference in BNNs is that the Bayes posterior can usually only be computed up to a constant factor.
This factor is called the \emph{marginal likelihood} or \emph{evidence} and is generally intractable to compute, since it involves a high-dimensional integral over the whole weight space.

\subsection{Stein Variational Gradient Descent}
\label{sec:svgd}


Stein Variational Gradient Descent (SVGD) \citep{liu2016stein} approximates the posterior of a BNN, while only needing access to its unnormalized posterior, and can thus be applied without computing the evidence.
It takes a set of $n$ particles $\{ \vw_1, \dots, \vw_n \}$ and evolves them according to a specific update rule, such that the particles converge in distribution to the target posterior \citep{liu2017stein, korba2020non}.
These particles can be used to approximate the predictive distribution~(Eq.~\ref{eqn:bayes_predictive}) by 
$
p(\vy^* | \vx^*, \vx, \vy) \approx \frac{1}{n} \sum_{i=1}^n p(\vy^* | f(\vx^*; \vw_i))
$.

The algorithm proceeds by choosing a positive-definite kernel function $k(\cdot, \cdot)$ in the particle space, sampling initial particles from some proposal distribution $q(\vw)$ as $\{ \vw^0_i | \vw^0_i \sim q(\vw) \}_{i=1}^n$, and then updating the particles according to the rule
\begin{equation}
    \vw_i^{t+1} \leftarrow \vw_i^t + \epsilon_t \phi (\vw_i^t) \; ; \; \phi(\vw_i) = \frac{1}{n} \sum_{j=1}^n \bigg[ \frac{1}{T} \underbrace{k(\vw_j,\vw_i) \nabla_{\vw_j} \log p(\vw_j|\vx, \vy)}_{\text{driving force}} + \underbrace{\nabla_{\vw_j} k(\vw_j,\vw_i)}_{\text{repulsive force}} \bigg].
    \label{eqn:update}
\end{equation}

Here, $\epsilon_t$ is the step size at time $t$ and $T$ is a temperature parameter to temper the posterior.
For $T=1$, this update follows a Wasserstein gradient flow of the KL divergence between the current particle distribution and the true Bayesian posterior $p(\vw|\vx, \vy)$ in the reproducing kernel Hilbert space (RKHS) defined by the kernel $k(\cdot, \cdot)$ \citep{liu2017stein}.

As can be seen in the update equation, the update direction for each particle consists of two parts: a driving force and a repulsive force.
The driving force moves the particles towards areas of higher posterior probability mass by following the gradient of the log posterior.
Crucially, the particles do not only follow their own gradient, but a mixture of all the other particles' gradients, weighted by the kernel.
In this way, the particles can inform each other about high-probability regions in the space.
The repulsive force, on the other hand, pushes the particles away from each other and therefore actively encourages diversity between the particles and avoids a collapse to a posterior mode.

SVGD approximates the posterior distribution by a set of $n$ particles, which can be viewed as an ensemble.
It has been shown that a neural network ensemble performs best if the members perform well individually and if they are diverse, that is, if they make complementary errors on held-out data \citep{fort2019deep, wenzel2020hyper}.
Although samples from the Bayes posterior naturally lead to a diverse ensemble in the asymptotic limit, we speculate that for a small sample size $n$, the repulsion in the SVGD might lead to a more diverse ensemble than one obtained by other approximate posterior sampling methods.



\subsection{Stochastic SVGD}
\label{sec:SVGD_sto}

Recently, \citet{gallego2018stochastic} proposed a stochastic variant of the SVGD algorithm that leads to the following system of discretized interacting stochastic differential equations (SDE): 

\begin{equation} 
\phi(\vw_i) = \frac{1}{n} \sum_{j=1}^n [k(\vw_j,\vw_i) \nabla_{\vw_j} \log \pi(\vw_j) + \nabla_{\vw_j} k(\vw_j,\vw_i)] + \sum_{j=1}^n \sqrt{\frac{2 \mathcal{K}_{ij}}{\epsilon_t}} \, \eta_j \, ,
\label{eqn:SVGD_stoc}
\end{equation}
with the matrix $\mathcal{K}_{ij} = \frac{1}{n}k(\vw_i,\vw_j) \mathbb{I}_{d \times d}$, and $\eta_j \sim \mathcal{N}(0,\mathbb{I}_{d \times d})$.


Given that the kernel gram matrix with elements $k(\vw_j,\vw_i)$ is positive definite, the SDE converges to the target distribution (see Appendix~\ref{app:sto_SVGD_conv}). The noise term in \eqref{eqn:SVGD_stoc} plays a key role, as it is designed to make the target invariant under the dynamics and thus ensures that the Markov chain evolving the single particles, \emph{even with a finite number of particles}, converges to $p(x)$ as $t \to \infty$.
In contrast, the deterministic case in \eqref{eqn:update} requires the mean field limit $n \to \infty$ for the convergence to be ensured. 
However, at a certain time step $t$, the stochastic SVGD might not yield the best \emph{finite} collection of particles to characterize the posterior, nor will their empirical measure $\rho_t(\vw)$ necessarily minimize the KL divergence with the posterior.
When compared to other sampling algorithms like Hamiltonian Monte Carlo (HMC) \citep{neal2012bayesian} or Stochastic gradient Langevin dynamics (SGLD) \citep{welling2011bayesian},
the stochastic SVGD has the advantage of naturally evolving multiple Markov chains, thus reducing the correlation between samples, and even more importantly, the repulsion between the chains allows for an efficient exploration of the posterior.

\section{Stein Variational Neural Network Ensembles} 
\label{sec:methods}


In the following, we present our unified framework for Stein variational neural network ensembles.
We study different kernel functions in the weight and function space, as well as deterministic and stochastic update rules.
This includes some existing approaches as special cases, such as standard deep ensembles \citep{lakshminarayanan2017simple}, BNN weight-space SVGD \citep{hu2019applying}, and BNN function-space SVGD \citep{wang2019function}, but it also includes several novel approaches.
We will lay out the motivation and theoretical properties for each approach and later proceed to empirically evaluating their respective performance.

Even though particle-based methods such as SVGD enjoy several guarantees in the asymptotic limit \citep{liu2017stein}, when using only a finite number of particles, there is empirical \citep{zhuo2018message} and theoretical \citep{zhang2020stochastic} evidence that particles in the original formulation of the SVGD method tend to collapse to a few local modes.
Moreover, in the BNN setting, the over-parameterized nature of BNN models leads to a target posterior which has a large number of modes that---despite their distance in the weight space---implement the same function \citep{badrinarayanan2015understanding}.
Predictive distributions approximated using samples from these modes do not improve over a simple point estimate and lead to a poor uncertainty estimation. This effect is also known in deep ensembles \citep{rahaman2020uncertainty}, where the absence of a constraint that prevents particles from converging to the same mode limits the possibility of improvement by introducing more ensemble members.
%
To overcome the above limitations of deep ensembles, as well as na\"ive SVGD, we investigate the different possibilities of incorporating the knowledge of the parameterized functions into the update rule of SVGD. To this end, we aim to use the repulsion in \eqref{eqn:update} (and \eqref{eqn:SVGD_stoc}) with a kernel acting in function space instead of the weight space to directly encourage functional diversity of the particles and avoid functional posterior collapse.

\subsection{Designing a functional kernel} 
\label{sec:f_kernel}

In this section, we show how to formulate a kernel which encapsulates the functional information of the different particles.
Let $g:(\vx,\vw) \mapsto f(\vx;\vw)$ be the map that maps a data point $\vx \in \mathcal{X}$ and weight vector $\vw$ to the corresponding neural network output.
We then denote the neural network outputs with weight $\vw_i$ as $\vf_i := f(\vx; \vw_i)$.
If we now define a kernel $k_f(\cdot, \cdot)$ over these function outputs (e.g., an RBF kernel), we can map it to a new kernel $k_w$ in the weight space as $k_w = k_f \circ g$.
That is, $k_w(\vw_i, \vw_j) = k_f(g(\vx, \vw_i), g(\vx, \vw_j))$. Note that this is a valid kernel due to the compositionality of kernels \citep[][eq. 6.19]{bishop2006pattern}.
Thus, we can use $k_w$ for weight-space inference using SVGD, while still taking functional diversity into account through the dependence on $k_f$.
In the following, we will generally drop the subscripts from $k_w$ and $k_f$ wherever it is obvious from the context which one is meant.

 It is important to notice that the map $g(\vx,\cdot)$ is not injective, due to the over-parameterization.
 This leads to a kernel that is only positive-semidefinite (not positive definite) and therefore there is no strict guarantee that the posterior distribution approximated with the particles will asymptotically converge to the true posterior in the weight space.
 However, each function still only corresponds to a finite number of weights \citep{badrinarayanan2015understanding} and the probability of the kernel matrix becoming singular at initialization is thus zero.
 Over the course of the optimization, the repulsive force in the update will keep the particles away from each other, such that they should also not converge to the same function and hence preserve the non-singularity of the matrix.
 At any rate, the main attention of our work is not on the quality of the posterior approximation in the weight space per se, but on an efficient characterization of the functional posterior distribution and the training of a functionally diverse ensemble.
 
For the SVGD methods, we focus on an RBF kernel in function space. Considering the infinite dimensionality of the function space, we use the expectation of the Euclidean norm in the RBF kernel, approximated with a Monte Carlo estimate over a batch of $B$ data points.
This yields
\begin{equation} 
\begin{split}
k^{RBF}_f(\vf_i, \vf_j) &= \exp \bigg( \mathbb{E}_{p(x)} \big[\left\lVert  f_i(\vx) - f_j(\vx)  \right\rVert^2 \big] \bigg)  
\approx \exp \bigg( \frac{1}{B} \left\lVert  f_i(\mX) - f_j(\mX)  \right\rVert^2 \bigg) \, ,
\end{split}
\end{equation}
where $f_i(\mX) = [f_i(\vx_1),...,f_i(\vx_B)]$ with $\vx \sim p(\vx)$, and $p(\vx)$ is the distribution of the data. Using this kernel leads to the repulsive force $ \frac{1}{n} \sum_{j=1}^n \nabla_{\vf_j} k(\vf_i,\vf_j) \propto  \frac{1}{n} \sum_{j\neq i} (\vf_i - \vf_j) k(\vf_i,\vf_j)$. 
 
\subsection{The functional SVGD update rule}
\label{sec:f_upd}


In the following, we illustrate different ways to define Stein neural network ensembles while taking the ensemble diversity into account.
In order to do so, we make use of the functional kernel introduced in the previous section to develop different update rules based on the original formulation in \eqref{eqn:update} and \eqref{eqn:SVGD_stoc} (we unify both equations via a parameter $\beta \in \{0,1\}$). We will consider the general case of maintaining $n$ particles in the parameter space and model their interaction with a general kernel $k(\cdot,\cdot)$ that might act in the function ($\vf$) or weight ($\vw$) space. We also consider the implicit functional likelihood $p(\vy|\vx, \vf)$, determined by the measure $p(\vy|\vx, \vw)$ in the weight space, as well as the functional prior $p(\vf)$, which can either be defined separately or modeled as a push-forward measure of the weight space prior $p(\vw)$.        
In the following, we discuss different choices of using the kernel in weight space or in function space and computing gradients with respect to weights or with respect to functions.
We discuss the advantages and disadvantages of the different choices and also include two novel approaches (fw-SVGD and h-SVGD).

\subsection{Weight-kernel weight SVGD (w-SVGD)}

When applied na\"ively to BNN inference, the original SVGD method introduced in Section~\ref{sec:svgd} (and its stochastic variant in Section~\ref{sec:SVGD_sto}) perform inference directly in the weight space, using a kernel that is also defined in the weight space \cite{hu2019applying}.
We denote this method by (w-SVGD) and show the full update rule here again to facilitate comparison with the alternative approaches:  
\begin{equation}
    \phi (\vw_i ) = \frac{1}{n} \sum_{j=1}^n \bigg( k(\vw_j,\vw_i) \nabla_{\vw_j} \log p(\vw_j| \vx,\vy)  + \nabla_{\vw_j} k(\vw_j,\vw_i)  \bigg) + \beta \sum_{j=1}^n \sqrt{\frac{2 \mathcal{K}_{ij}}{ \epsilon_t}} \, \eta_j
    \label{eqn:ww_update}
\end{equation}
with  $\mathcal{K}_{ij} = \frac{1}{n}k(\vw_i,\vw_j) \mathbb{I}_{d \times d}$ and $\eta_j \sim \mathcal{N}(0,\mathbb{I}_{d \times d})$. This update rule in its deterministic ($\beta = 0$) and stochastic ($\beta = 1$) version encourages diversity between weights, but---due to the overparameterization mentioned above \cite{badrinarayanan2015understanding}---not necessarily between functions.

\subsection{SVGD as Deep ensembles (DE) and SGLD}

When using the indicator function as a kernel scaled by the number of particles, that is $k(\vw_i,\vw_j) = n \, \indicator(\vw_i,\vw_j)$, where $\indicator(\vw_i,\vw_j)$ is unity if the two arguments are equal and zero otherwise, the w-SVGD simplifies to the standard deep ensemble, reducing for $\beta = 0$ to MAP estimations:
\begin{equation}
\begin{split}
    \phi (\vw_i )
    &= \nabla_{\vw_i} \log p(\vw_i| \vx,\vy) + \beta \sqrt{\frac{2}{ \epsilon_t}} \, \eta_j
    \label{eqn:de_update}
\end{split}
\end{equation}

We can see immediately that this removes any repulsion and thus does not encourage diversity in the ensemble any more, neither in the weight nor the function space. Moreover, for $\beta = 1$, we recover the update rule for the SGLD algorithm \cite{welling2011bayesian}.

\subsection{Function-kernel weight SVGD (fw-SVGD)} 

The simplest solution to take functional diversity into account and operate on functions instead of the weights, is to use the functional kernel, as illustrated in Section~\ref{sec:f_kernel}, and substitute it in \eqref{eqn:ww_update}:
\begin{align}
    \phi (\vw_i )
         &= \frac{1}{n} \sum_{j=1}^n \bigg( k(\vf_i,\vf_j) \bigg(\frac{\partial \vf_j}{\partial \vw_j}\bigg)^\top \frac{\partial \log p(\vf_j| \vx,\vy)}{\partial \vf_j} +  \bigg(\frac{\partial \vf_j}{\partial \vw_j} \bigg)^\top  \frac{\partial k(\vf_i,\vf_j)}{\partial \vf_j} \bigg) + \beta \sum_{j=1}^n \sqrt{\frac{2 \mathcal{K}_{ij}}{ \epsilon_t}} \, \eta_j \nonumber\\
         &= \frac{1}{n} \sum_{j=1}^n \mJ_j^\top \bigg( k(\vf_i,\vf_j)  \nabla_{\vf_j} \log p(\vf_j| \vx,\vy) + \nabla_{\vf_j} k(\vf_i,\vf_j) \bigg) + \beta \sum_{j=1}^n \sqrt{\frac{2 \mathcal{K}_{ij}}{ \epsilon_t}} \, \eta_j
\label{eqn:fw_update}
\end{align}
where $\mathcal{K}_{ij} = \frac{1}{n}k(\vf_i,\vf_j) \mathbb{I}_{d \times d}$ and  $\mJ_j = \partial \vf_j / \partial \vw_j$ denotes the Jacobian of $\vf_j$ with respect to $\vw_j$. Using the chain rule, we could isolate the factor corresponding to the exact SVGD performed in function space.
Moreover, we can interpret the role of the Jacobian $\mJ_j^\top$ for each particle as a projector that maps the gradients from the function space into the weight space.

Intuitively, the main issue with this approach is that the functional kernel might be a good measure for functional diversity, but a bad similarity measure for averaging the log-posterior gradients in the weight space.
As mentioned before, similar functions might be implemented by very different weights and may therefore share gradients in this approach even though they are far away from each other in the weight space.
However, the method can still satisfy convergence under some assumptions: For the deterministic case ($\beta = 0$), the minimization of the KL divergence can be ensured provided that the functional kernel introduced in Section~\ref{sec:f_kernel} belongs to the Stein class. For the stochastic case ($\beta = 1$), a weaker assumption is sufficient to have convergence to the posterior in the asymptotic limit, namely that the kernel gram matrix of the functional kernel is positive definite at each step, such that the result in Sec.~\ref{sec:SVGD_sto} holds (see Appendix~\ref{app:sto_SVGD_conv}, Theorem~\ref{the:convergence_s_SVGD}).
As already mentioned in Section~\ref{sec:f_kernel} the functional kernel is not positive definite in general due to the overparametrization issue, but, in the finite particle regime, the repulsion ensures that the gram matrix is non-singular at every step.

\subsection{Functional SVGD (f-SVGD)}
\label{sec:f-svgd}
    
A similar version of the previous update rule was introduced by \citet{wang2019function} :
\begin{equation} 
\begin{split} 
    \phi (\vw_i )
     &= \frac{1}{n} \sum_{j=1}^n \mJ_i^\top \bigg( k(\vf_i,\vf_j)  \nabla_{\vf_j} \log p(\vf_j| \vx,\vy) +  \nabla_{\vf_j} k(\vf_i,\vf_j) \bigg) \, .
\end{split} 
\label{eqn:f_update}
\end{equation}
One major issue with this approach, compared to the previous ones, is the requirement of a function space prior gradient: $  \nabla_{\vf_j} \log p (\vf_j|\vx,\vy) = \nabla_{\vf_j} \log p (\vy|\vx,\vf_j) + \nabla_{\vf_j} \log p (\vf_j) $. 
If one wants to use the same prior as for the other approaches, namely a standard BNN prior defined in weight space, an additional estimator is needed \citep{sun2019functional}, since the implicit prior in function space defined by the one in weight space is not analytically accessible. In this work, we adopted the spectral gradient estimator for implicit distributions (SSGE), introduced by \citet{shi2018spectral}.
Alternatively, a separate prior directly in the function space could be defined, for instance, using a Gaussian process (GP).
This would however exacerbate the comparison with the other presented approaches and can also lead to optimization issues due to pathologies when comparing GP posteriors and BNN posteriors in function space \cite{burt2020understanding}.

Another important observation is the difference between the  update rules given by \eqref{eqn:fw_update} and \eqref{eqn:f_update} in the used Jacobian matrix: crucially, \eqref{eqn:f_update} only uses the Jacobian matrix of particle $i$ which we aim to update. This difference is fundamental and leads to completely different inference dynamics, since in the f-SVGD update, all the gradients are computed in function space and are then projected back using exclusively the $i$-th Jacobian. In contrast, the fw-SVGD update in \eqref{eqn:fw_update} uses a separate Jacobian $\mJ_j$ for each gradient $\nabla_{\vf_j}$ and thus calculates an expectation in the weight space instead of the function space.
Intuitively, this means that the f-SVGD approximates the map $g$ from weight to function space with one global linear function, extrapolated from the single point $\vf_i$, while the fw-SVGD approximates $g$ locally at each point $\vf_j$ with $\mJ_j$.
The f-SVGD's global linearization is problematic, since we have strong reasons to believe that this map is very complex and highly nonlinear in reality \cite{badrinarayanan2015understanding}.
The projected gradient $\mJ_i^\top \nabla_{\vf_j}$ can thus differ quite significantly from the true gradient $\mJ_j^\top \nabla_{\vf_j}$ which is used by the fw-SVGD.
We would therefore expect the averaged gradients provided by fw-SVGD to be more informative than the ones from f-SVGD.

\subsection{Hybrid-kernel weight SVGD (h-SVGD)}

Finally, we consider another novel update rule that combines a kernel in the weight space, to correctly average the posterior gradient contributions of the different particles, with a functional kernel, to repel particles parameterizing the same function and thus encourage functional diversity:
\begin{equation}
\begin{split}
    \phi (\vw_i )= \frac{1}{n} \sum_{j=1}^n \bigg( k(\vw_i,\vw_j) &\nabla_{\vw_j} \log p(\vw_j| \vx,\vy) + \nabla_{\vw_j} k(\vf_i,\vf_j))  \bigg) + \beta \sum_{j=1}^n \sqrt{\frac{2 \mathcal{K}_{ij}}{ \epsilon_t}} \, \eta_j
\end{split}
\label{eqn:mw_update}
\end{equation}
Since this update rule uses two different kernels, it does not technically constitute a version of the original SVGD, and thus does not inherit its strong convergence guarantees. Moreover, for the stochastic version ($\beta =1$), it is not possible to guarantee that the stationary distribution is uniquely the posterior.
However, in practice, we find that this constitutes a useful tradeoff between weight-space and function-space methods and yields ensemble members with performant functions, while still encouraging functional diversity. We leave a deeper theoretical analysis of this SVGD scheme for future work and present our empirical findings in the following.




\section{Experiments}
\label{sec:experiments}

We evaluate the different SVGD methods described above and compare them against deep ensembles and HMC. 
First, we present a qualitative evaluation on a synthetic regression and a  classification task.
We then evaluate the predictive performance on standard real-world classification datasets and assess the uncertainty estimation of the methods in terms of calibration and out-of-distribution (OOD) detection.  
For the HMC, we initialize multiple parallel chains to reduce autocorrelation between the samples.
In our experiments, we report the test accuracy, negative log-likelihood (NLL), and the expected calibration error (ECE) \cite{naeini2015obtaining}. To assert the robustness on out-of-distribution (OOD) data, we report the ratio between predictive entropy on OOD and test data points ($H_{o}/H_{t}$), and the OOD detection area under the ROC curve AUROC(H) \cite{liang2017enhancing}. Moreover, to assess the  diversity of the ensemble generated by the different methods in function space, we measure the functional diversity as illustrated in Appendix~\ref{sec:model_dis}.
In particular, we report the ratio between the average model disagreement on the OOD and test data points ($MD_{o}/MD_{t}$) and additionally the AUROC(MD) computed using this measure instead of the entropy. 
Additional experimental results are deferred to Appendix~\ref{sec:additional_exp} and implementation details, including hyperparameters, to Appendix~\ref{sec:imp_details}. In all experiments, the letter \textbf{-l} added next to a method indicates that the functional kernel is evaluated on the logits (i.e., pre-softmax activations), whereas the letter \textbf{-s} indicates the evaluation on the softmax outputs (i.e., probabilities).

\paragraph{Low-dimensional synthetic problems}

We first assess the methods on a synthetically generated one-dimensional regression task and a two-dimensional classification task,
the results are reported in Figure~\ref{fig:toy_tasks}.
We can see that in the regression experiment (top row) all the methods that work exclusively in the weight space (DE, w-SVGD) are unable to capture the epistemic uncertainty between the two clusters of the data points.
Conversely, the functional methods (f-SVGD, fw-SVGD, h-SVGD) are able to capture the functional diversity in this region leading to a predictive posterior distribution that closely resembles the gold standard generated with HMC.




Similarly, in the classification setting, we observe that the weight-space methods (DE, w-SVGD) are overconfident and do not capture the uncertainty well. These methods only account for uncertainty close to the decision boundaries. In contrast, the functional methods are confident close to the true data and capture the increase in epistemic uncertainty away from the data points.
These observations transfer to the setting with stochastic updates (see Fig.~\ref{fig:toy_tasks_sto} in the Appendix), but the stochastic methods all tend to capture the uncertainty better than their deterministic counterparts.



    
\paragraph{Image classification on FashionMNIST}

Next, we test the methods in an image classification setting using the FashionMNIST dataset \cite{xiao2017fashion} for training and the MNIST dataset \cite{lecun1998mnist} as out-of-distribution (OOD) data.
We see in Table~\ref{tab:fmnist} (top) that h-SVGD-l outperforms the other methods in terms of accuracy, NLL, OOD detection using MD, functional diversity, and calibration.
The f-SVGD instead achieves the best OOD AUROC in entropy, but a weak test accuracy.
Moreover, the model disagreement ratio shows that most of the methods acting in function space have higher functional diversity when compared with standard deep ensembles, thus confirming our intuition.
This suggests that, in contrast to its lack of theoretical guarantees, the h-SVGD may indeed constitute a good balance between weight-space and function-space behavior and could offer a practical method for training diverse neural network ensembles.

\begin{table*}[t]
\caption{\textbf{FashionMNIST classification.} \textbf{AUROC(H)} is the AUROC computed using the entropy whereas \textbf{AUROC(MD)} is computed using the model disagreement. $\mathbf{H_{o}/H_{t}}$ is the ratio of the entropies on OOD and test points, respectively, and $\mathbf{MD_{o}/MD_{t}}$ is the ratio for model disagreement.
$\beta = 0$: We see that the h-SVGD-l performs best in terms of accuracy, NLL, OOD detection using MD and leads to the most functionally diverse and calibrated ensemble. The f-SVGD instead achieves the best OOD AUROC in entropy, but a weak test accuracy.
$\beta = 1$: The fw-SVGD-l achieves the best accuracy, but the fw-SVGD-s yields the best OOD detection using both entropy and MD.}
\centering
\begin{adjustbox}{max width=\textwidth}
\begin{tabular}{cllllllll}
\toprule
&& \textbf{AUROC(H)}   &  \textbf{AUROC(MD)}  & \textbf{Accuracy}            & $\mathbf{H_{o}/H_{t}}$   & $\mathbf{MD_{o}/MD_{t}}$ & \textbf{ECE}       & \textbf{NLL}  \\ 
\midrule

\multirow{7}{*}{\rotatebox[origin=c]{90}{$\beta = 0$}} &\textbf{DE}  &  0.938$\pm$0.001 &    0.977$\pm$0.001 &  88.864$\pm$0.015 &  4.445$\pm$0.013 &  7.566$\pm$0.011 &    0.013$\pm$0.001 &    0.179$\pm$0.001 \\
&\textbf{w-SVGD} &  0.941$\pm$0.001 &  0.978$\pm$0.001 &   88.280$\pm$0.041 &   4.299$\pm$0.020 &   7.422$\pm$0.010 &  0.017$\pm$0.001 &  0.191$\pm$0.001 \\
&\textbf{f-SVGD} &  \textbf{0.974$\pm$0.001} &  0.979$\pm$0.001 &   87.780$\pm$0.101 &   3.220$\pm$0.031 &  7.767$\pm$0.023 &  0.067$\pm$0.001 &  0.262$\pm$0.003 \\
&\textbf{fw-SVGD-l}  &  0.948$\pm$0.001 &  0.979$\pm$0.001 &   88.792$\pm$0.020 &   4.342$\pm$0.010 &  7.303$\pm$0.004 &    0.021$\pm$0.001 &     0.190$\pm$0.001 \\
&\textbf{fw-SVGD-s} &    0.958$\pm$0.001 &    \textbf{0.986$\pm$0.001} &   88.510$\pm$0.034 &  4.713$\pm$0.024 &  7.783$\pm$0.036 &   \textbf{0.010$\pm$0.001} &  0.182$\pm$0.001 \\
 &\textbf{h-SVGD-l}  &  0.959$\pm$0.001 &  \textbf{0.985$\pm$0.001} &   \textbf{89.080$\pm$0.034} &  \textbf{4.933$\pm$0.007} &  \textbf{7.835$\pm$0.018} &    \textbf{0.011$\pm$0.001} &    \textbf{0.173$\pm$0.001} \\
&\textbf{h-SVGD-s}  &  0.931$\pm$0.001 &  0.969$\pm$0.001 &   88.870$\pm$0.015 &  4.356$\pm$0.007 &  7.576$\pm$0.019 &    0.016$\pm$0.001 &    0.183$\pm$0.001 \\

\midrule 

\multirow{6}{*}{\rotatebox[origin=c]{90}{$\beta = 1$}} & \textbf{pSGLD} &  0.902$\pm$0.001 &  0.957$\pm$0.001 &  88.562$\pm$0.049 &  3.767$\pm$0.015 &  6.974$\pm$0.016 &  0.022$\pm$0.001 &  0.194$\pm$0.001 \\
& \textbf{w-SVGD} &  0.925$\pm$0.001 &    0.971$\pm$0.001 &   87.390$\pm$0.032 &   3.868$\pm$0.020 &  7.335$\pm$0.027 &  0.018$\pm$0.001 &  0.208$\pm$0.001 \\
& \textbf{fw-SVGD-l} &  0.951$\pm$0.001 &  0.982$\pm$0.001 &  \textbf{88.876$\pm$0.035} &   4.370$\pm$0.011 &   7.442$\pm$0.020 &  0.021$\pm$0.001 &   0.190$\pm$0.001 \\
& \textbf{fw-SVGD-s} &  \textbf{0.955$\pm$0.001} &  \textbf{0.984$\pm$0.001} &  88.616$\pm$0.027 &   \textbf{4.653$\pm$0.010} &  \textbf{7.718$\pm$0.021} &   \textbf{0.013$\pm$0.001} &    \textbf{0.183$\pm$0.001} \\
& \textbf{h-SVGD-l}  &  0.929$\pm$0.001 &    0.974$\pm$0.001 &  87.524$\pm$0.036 &  3.836$\pm$0.013 &  7.253$\pm$0.031 &  0.023$\pm$0.001 &  0.211$\pm$0.001 \\
& \textbf{h-SVGD-s}  &  0.933$\pm$0.001 &    0.975$\pm$0.001 &   87.730$\pm$0.031 &  3.877$\pm$0.018 &   7.393$\pm$0.010 &  0.025$\pm$0.001 &   0.210$\pm$0.001 \\

\bottomrule

\end{tabular}   
\end{adjustbox}
\label{tab:fmnist}
\end{table*}

In Table~\ref{tab:fmnist} (bottom), we also see the results for the stochastic updates.
In this case, the fw-SVGD-l is performing best in terms of accuracy whereas fw-SVGD-s achieves the best OOD detection, calibration, and functional diversity.
Interestingly, the stochastic updates improve the test accuracy when the covariance matrix is based on the function space kernel (f-SVGD, fw-SVGD-l, fw-SVGD-s), whereas for the weight space methods performance seem to degenerate. Moreover, we observe that computing the AUROC with the model disagreement (MD) leads to higher performance in comparison to the entropy across all methods.
These observations are also qualitatively reflected in the calibration curves shown in Appendix~\ref{sec:calibration_curves} (Fig.~\ref{fig:cal_mnist}) and corruption analysis in Appendix~\ref{sec:corruption}.

\paragraph{Image classification on CIFAR-10} 

In this experiment, we use a residual network (ResNet32) architecture \cite{he2016deep} on CIFAR-10 \cite{krizhevsky2009learning}.
Here, we use the SVHN dataset \cite{netzer2011reading} as OOD data.
We see in Table~\ref{tab:cifar} (top) that the deep ensemble is outperforming all other methods for the deterministic case in terms of accuracy and negative log likelihood. However, the h-SVGD-s performs better in terms of OOD detection using the AUROC with the entropies and achieves a better functional diversity as measured by the ratio of entropies and MD. Both methods appear to be similarly calibrated. 
 
Interestingly, the f-SVGD behaves very differently on this dataset than on FashionMNIST. In this case, not only is its accuracy the lowest, but also the OOD detection is clearly underperforming when compared to the other methods.
This is probably due to the fact that the mapping from weights to functions is even more nonlinear for these large networks and that the linearization with the local Jacobian (see Sec.~\ref{sec:f-svgd}) is therefore an even worse approximation, leading to worse results in the weight space.
The fw-SVGD, on the other hand, uses the correct local Jacobian for each functional gradient and does not suffer from this issue.

The stochastic case is also shown in Table~\ref{tab:cifar} (bottom). Here, the introduced stochasticity leads to better results overall when compared to the deterministic cases in terms of both accuracy and OOD detection. Specifically, the  h-SVGD-s, due to its enhanced functional diversity, leads to the best OOD performance and NLL. In comparison, the same method applied on the logits (h-SVGD-l) gives the best accuracy, showing an improvement of 1.2~\% over pSGLD. These results are also confirmed in the calibration curves reported in Appendix~\ref{sec:calibration_curves} (Figure~\ref{fig:cal_cifar}) where the h-SVGD-l has the largest area under the curve. 
Note that we do not use data augmentation nor batch normalization here, such that the achieved 85~\% accuracy comes close to the state of the art for BNNs on this dataset \cite{fortuin2020bayesian}.

\begin{table*}
\caption{\textbf{CIFAR-10 classification.} $\beta = 0$: The deep ensemble performs best in terms of accuracy and likelihood, while the h-SVGD-s offers better OOD detection and functional diversity. The f-SVGD instead shows the worst results in terms of both OOD detection and accuracy. $\beta = 1$: The h-SVGD-l performs best in terms of accuracy, while the h-SVGD-s offers better OOD detection and functional diversity. However, the pSGLD achieves the best calibration (ECE).}
\centering
\begin{adjustbox}{max width=\textwidth}
\begin{tabular}{clllllll}
\toprule
&& \textbf{AUROC}(H) & \textbf{Accuracy} & $\mathbf{H_{o}/H_{t}}$ & $\mathbf{MD_{o}/MD_{t}}$ & \textbf{ECE} & \textbf{NLL}  \\ 
\midrule
\multirow{6}{*}{\rotatebox[origin=c]{90}{$\beta = 0$}} &\textbf{DE}                         &  0.854$\pm$0.004 &   \textbf{85.434$\pm$0.067} &    2.199$\pm$0.010 &  1.527$\pm$0.012 &   \textbf{0.064$\pm$0.001} &     \textbf{0.296$\pm$0.002}        \\
&\textbf{w-SVGD}                     &  0.838$\pm$0.004 &  85.338$\pm$0.048 &  2.104$\pm$0.017 &  1.478$\pm$0.011 &  0.065$\pm$0.001 &    0.299$\pm$0.001        \\  
&\textbf{f-SVGD}                     &  0.782$\pm$0.002 &  81.948$\pm$0.041 &   1.730$\pm$0.003 &  1.467$\pm$0.001 &  0.208$\pm$0.003 &  0.311$\pm$0.001       \\
&\textbf{fw-SVGD-l} &  0.852$\pm$0.003  &   85.234$\pm$0.040 &  2.136$\pm$0.004 &  1.459$\pm$0.007 &  0.071$\pm$0.001 &    0.307$\pm$0.001         \\
&\textbf{fw-SVGD-s}  &  0.843$\pm$0.002 &  85.364$\pm$0.018 &  2.097$\pm$0.015 &  1.409$\pm$0.012 &   0.070$\pm$0.001 &    0.304$\pm$0.001          \\
&\textbf{h-SVGD-l}  &  0.853$\pm$0.002  &   85.230$\pm$0.098 &  2.189$\pm$0.007 &  1.481$\pm$0.007 &  \textbf{0.064$\pm$0.001} &      0.300$\pm$0.002           \\
&\textbf{h-SVGD-s}  &   \textbf{0.861$\pm$0.002}  &   85.250$\pm$0.049 &  \textbf{2.224$\pm$0.003} &   \textbf{1.537$\pm$0.004} &  0.069$\pm$0.008 &        0.303$\pm$0.001 \\

\midrule
\multirow{6}{*}{\rotatebox[origin=c]{90}{$\beta = 1$}} & \textbf{pSGLD}                & 0.861$\pm$0.002 &	85.818$\pm$0.035    &	2.253$\pm$0.007  &	1.517$\pm$0.005     &	\textbf{0.065$\pm$0.001}     &	0.290$\pm$0.001   \\
&\textbf{w-SVGD}                 &  0.878$\pm$0.002            &    86.720$\pm$0.073           &       2.452$\pm$0.004            &      1.540$\pm$0.001      &  0.223$\pm0.004$           &  0.275$\pm$0.002 \\
&\textbf{fw-SVGD-l}   &  0.844$\pm$0.005           &    85.754$\pm$0.079           &       2.109$\pm$0.008             &      1.448$\pm$0.005      &  0.254$\pm$0.008          &  0.306$\pm$0.001 \\
&\textbf{fw-SVGD-s} &  0.852$\pm$0.003           &    85.284$\pm$0.096          &       2.112$\pm$0.012           &      1.407$\pm$0.006      &  0.246$\pm$0.009         &  0.308$\pm$0.002\\
&\textbf{h-SVGD-l}    &  0.880$\pm$0.001      & \textbf{86.862$\pm$0.056}     &       2.454$\pm$0.016             &      1.580$\pm$0.001      &  0.201$\pm$0.003  &  0.270$\pm$0.001\\
&\textbf{h-SVGD-s}  &  \textbf{0.890$\pm$0.002}   &    86.666$\pm$0.037          &\textbf{2.521$\pm$0.006}           &\textbf{1.600$\pm$0.002}    &  0.220$\pm$0.006          &  \textbf{0.269$\pm$0.001} \\

\bottomrule

\end{tabular}   
\end{adjustbox}
\label{tab:cifar}
\end{table*}

\section{Related Work}
\label{sec:related_work}
We have already addressed some of the related work in Section~\ref{sec:background} and additionally highlight previous relevant work on ensemble-based inference methods for deep neural networks here.

\smallskip\noindent\textbf{Particle-based inference.}
Particle-based methods for flexible inference have gained a lot of momentum since the publication of the original SVGD method (w-SVGD) \citep{liu2016stein}.
Out of these, standard SVGD is still the best studied method, with many theoretical convergence results and strong guarantees \citep{liu2017stein, liu2019understanding, korba2020non, duncan2019geometry, chewi2020svgd}.
Interestingly, it has also been shown that SVGD performance can benefit from projecting the particles into a different space \citep{chen2020projected}, which is very similar to our projection from weight space into function space.
Moreover, our approximation of the function space Stein discrepancy using mini-batches is related to the idea of stochastic Stein discrepancies, which have been shown to offer good approximations in practice \citep{gorham2020stochastic}.
Apart from that, many extensions of SVGD have been proposed, including using second-order methods \citep{detommaso2018stein}, additional Langevin noise \citep{gallego2018stochastic}, importance-weighting \citep{yao2020stacking}, non-Markovian trajectories \citep{ye2020stein}, explicit noise models \citep{chang2020kernel}, and matrix-valued kernels \citep{wang2019stein}.
While we study standard SVGD in this work, most of these extensions could be readily applied in our framework, which will be an exciting avenue for future work.

\smallskip\noindent\textbf{BNN inference.}
Bayesian neural networks have a long history, dating back to \citet{mackay1992practical} and \citet{neal1995bayesian}.
While Markov Chain Monte Carlo (MCMC) inference is considered the gold standard for these models \citep{neal1995bayesian, wenzel2020good, fortuin2020bayesian, garriga2021exact, fortuin2021bnnpriors, fortuin2021priors, izmailov2021bayesian}, it is often costly and in practice replaced by variational inference (VI) approaches \citep{blundell2015weight, swiatkowski2020k, dusenberry2020efficient, immer2021scalable}.
So far, SVGD has been applied to BNNs only sparingly \citep{hu2019applying, wang2019function}. These approaches correspond to special instantiations of our framework. 
Moreover, many practical techniques for improving BNN performance can also be applied in our framework: posterior tempering \citep{mandt2016variational, wenzel2020good} corresponds to the temperature factor in \eqref{eqn:update}, and the cyclical annealing schedules from VI \citep{fu2019cyclical} and MCMC inference \citep{zhang2019cyclical} are very related to the annealed SVGD \citep{dangelo2021annealed}.

 
\section{Conclusion}

We have presented a unified framework for ensembling methods for deep neural networks, including many popular methods as special cases, such as deep ensembles and standard SVGD.
We have shown that incorporating functional repulsion between ensemble members is non-trivial, but that it can significantly improve the quality of the estimated uncertainties and OOD detection on synthetic and real-world data and can approach the true Bayesian posterior more closely.
Moreover, we have shown that stochastic SVGD updates can outperform the conventional deterministic ones and that some previously undescribed SVGD variants (fw-SVGD and h-SVGD) can outperform the standard methods on some tasks.

\bibliographystyle{plainnat}
\bibliography{references}

\newpage

\onecolumn

\appendix
\counterwithin{figure}{section}
\counterwithin{table}{section}
\section*{\Large{Supplementary Material: On Stein Variational Neural Network Ensembles}}

\section{Additional definitions}
\subsection{Quantify functional diversity}
\label{sec:model_dis}
As illustrated in Section~\ref{sec:bnns} , in the Bayesian context, predictions are made by doing a Monte-Carlo estimation of the BMA. Functional diversity, and so the diversity in the hypotheses taken in consideration when performing the estimation, determines the epistemic uncertainty and the confidence over the predictions. Importantly, the epistemic uncertainty allows for the quantification of the likelihood of a test point to belong to the same distribution from which the training data points where sampled \cite{ovadia2019can}. Following this, the uncertainty can be used for the problem of Out-of-distribution (OOD) detection \cite{chen2020robust} that is often linked to the ability of a model to \emph{”know what it doesn't know”}. A common way used in the literature to quantify the uncertainty is the  Entropy\footnote{The continuous case is analogous using the differential entropy} $\mathcal{H}$ of the predictive distribution: 
\begin{equation}
    \mathcal{H} \big\{  p(\mb{y}'| \mb{x}', \mathcal{D})  \big\} = - \sum_y  p(\mb{y}'| \mb{x}', \mathcal{D}) \log  p(\mb{y}'| \mb{x}', \mathcal{D}) \, .
\end{equation}
Nevertheless, it has been argued in recent works \cite{malinin2019ensemble} that this is not a good measure of uncertainty because it does not allow for a disentanglement of epistemic and aletoric uncertainty. Intuitively, we would like the predictive distribution of an OOD point to be uniform over the different classes. However, using the entropy and so the average prediction in the BMA, we are not able to distinguish between the case in which all the hypotheses disagree very confidently due to the epistemic uncertainty or are equally not confident due to the aleatoric uncertainty. To overcome this limitation, we can use a direct measure of the model disagreement computed as: 
\begin{equation}
    \mathcal{MD}^2(\mb{y}';\mb{x}',\mathcal{D}) = \int_\vw \big[ p(\mb{y}'| \mb{x}', \vw)-p(\mb{y}'| \mb{x}', \mathcal{D})\big]^2 p(\vw| \mathcal{D}) d \vw \, .
    \label{eqn:model_dis}
\end{equation}
It is easy to see how the quantity in \eqref{eqn:model_dis}, measuring the deviation from the average prediction is zero when all models agree on the prediction. The latter can be the case of a training point where all hypotheses are confident or a noisy point where all models \emph{”don't know”} the class and are equally uncertain. On the other side the model disagreement will be greater the zero the more the model disagree on a prediction representing like this the epistemic uncertainty. To obtain a scalar quantity out of \eqref{eqn:model_dis} we can consider the expectation over the output space of $\mb{y}$: 
\begin{equation}
    \mathcal{MD}^2(\mb{x}') = \mathbb{E}_{y} \bigg[\int_\vw \big[ p(\mb{y}'| \mb{x}', \vw)-p(\mb{y}'| \mb{x}', \mathcal{D})\big]^2 p(\vw| \mathcal{D}) d \vw \bigg] \, .
    \label{eqn:model_dis_norm}
\end{equation}

\subsection{Stochastic SVGD convergence}
\label{app:sto_SVGD_conv}
Interestingly, \citep{gallego2018stochastic} noticed how the stochastic update rule in \eqref{eqn:SVGD_stoc} can be framed in the framework developed in \cite{ma2015complete} by considering an augmented space given by the concatenation of the particles $\tilde{x} = (x_1,\dots, x_n) \in \mathbb{R}^{nd}$ to obtain a valid MCMC-type posterior sampler that naturally runs $n$ interacting Markov chains. With this design the method falls into the class of settings for which Theorem 1 in \citet{ma2015complete} is valid; with the condition that the matrix $\mathbf{D}$, and so the gram matrix of the kernel in this case, is positive definite: 
 \begin{theorem}[Convergence of stochastic SVGD \cite{gallego2018stochastic}]
     Assuming a positive definite kernel gram matrix with elements $k(\vw_j,\vw_i)$, the discretization of the stochastic SVGD SDE in \eqref{eqn:SVGD_stoc} has the product measure of the target $\tilde{p}(\tilde{\vw}) = \prod_{j=1}^n p(\vw_j)$ as a unique stationary distribution. Convergence is ensured in the asymptotic limit of the step size $\epsilon_t \to 0$ under the Robbins-Monro conditions.
 \label{the:convergence_s_SVGD}
 \end{theorem}

\section{Additional experimental results}
\label{sec:additional_exp}

In this section, we show the results for the stochastic versions on the synthetic data and complementary results for FashionMNIST and CIFAR-10.

\subsection{Low-dimensional synthetic problems}
\begin{figure*}[h]
\centering
    \begin{subfigure}[b]{0.18\textwidth}
    \includegraphics[width=\linewidth,trim={1cm 1cm 1cm 0cm},clip]{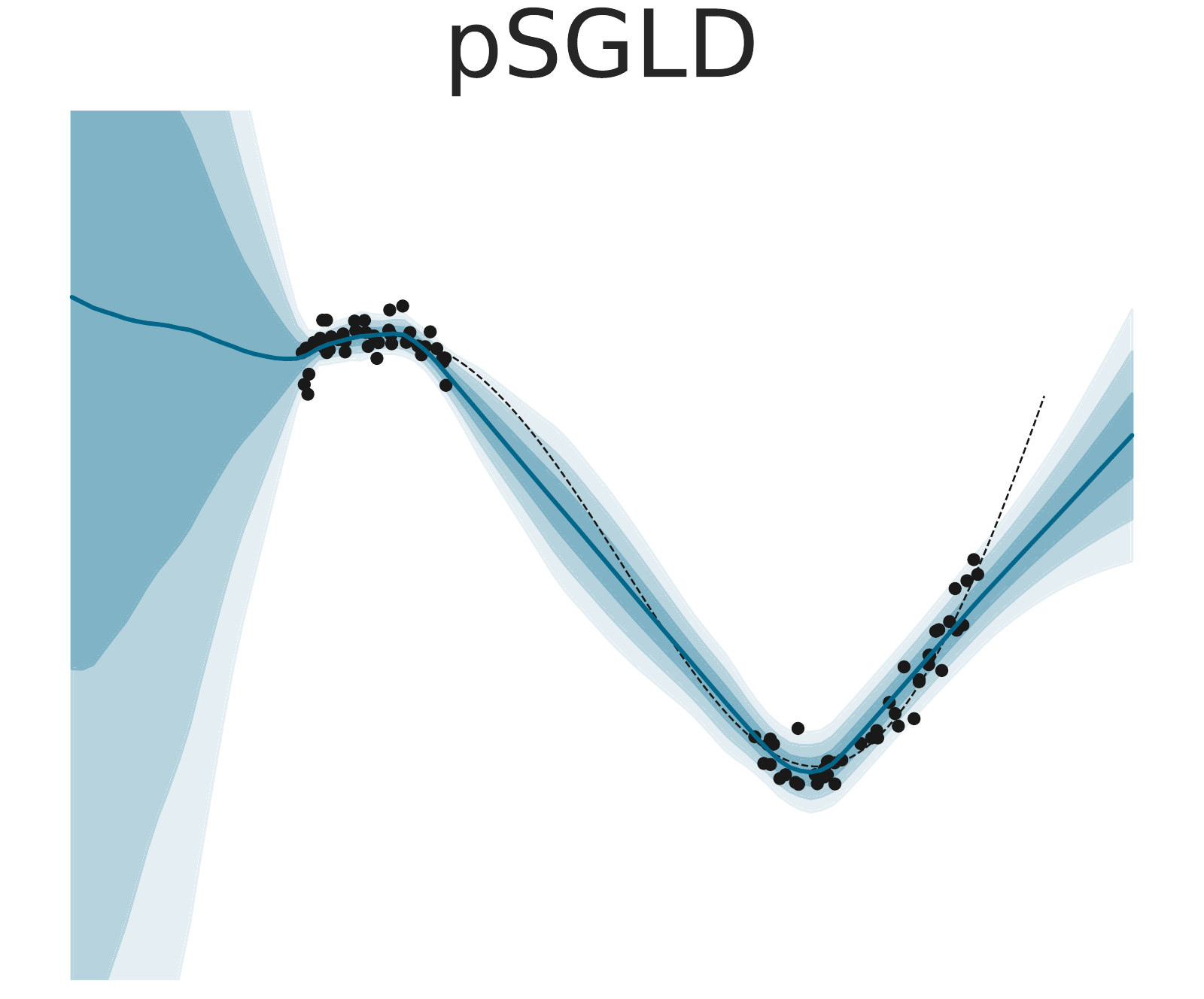}
    \end{subfigure}
    \begin{subfigure}[b]{0.18\textwidth}
    \includegraphics[width=\linewidth,trim={1cm 1cm 1cm 0cm},clip]{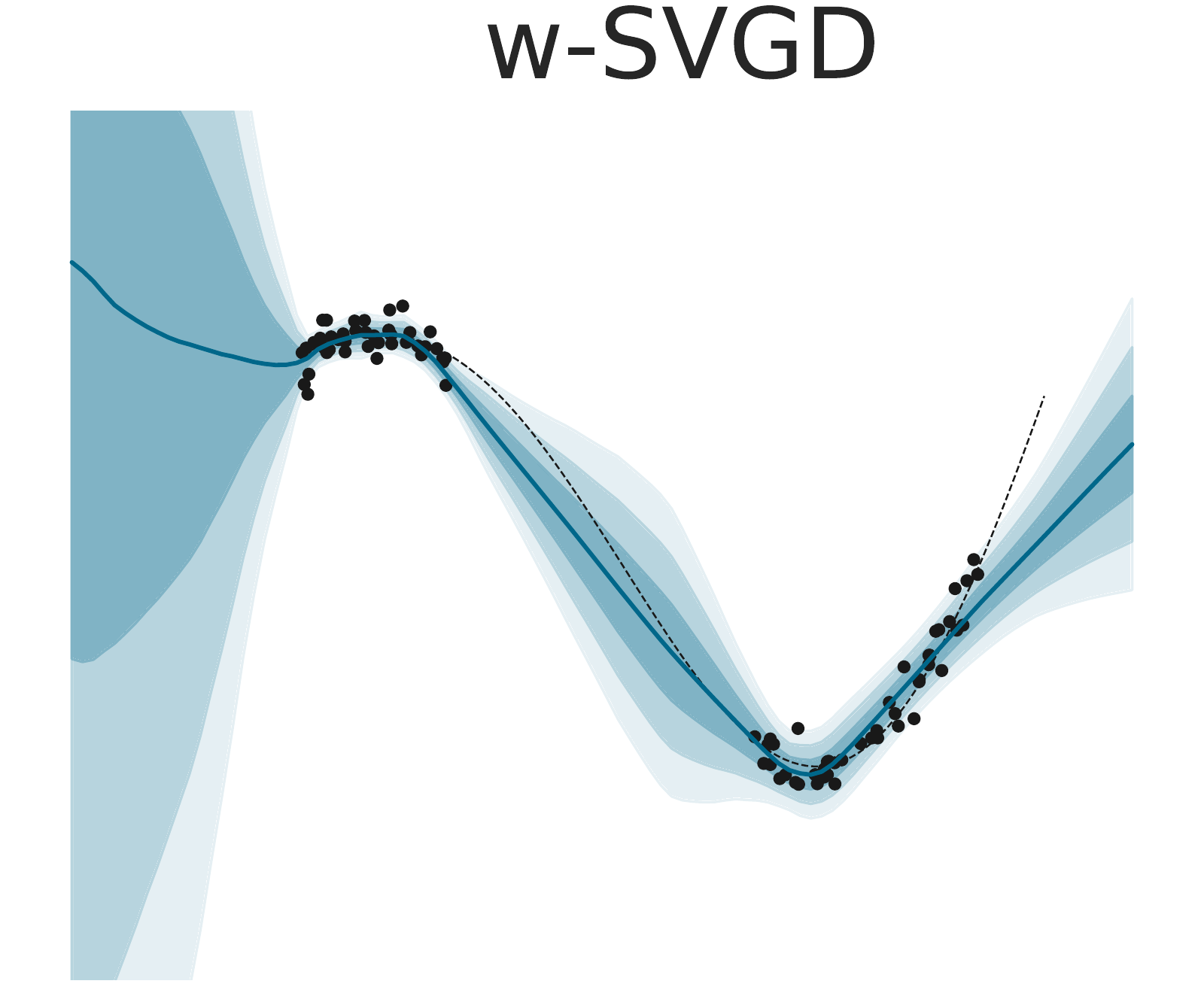}
    \end{subfigure}
    \begin{subfigure}[b]{0.18\textwidth}
    \includegraphics[width=\linewidth,trim={1cm 1cm 1cm 0cm},clip]{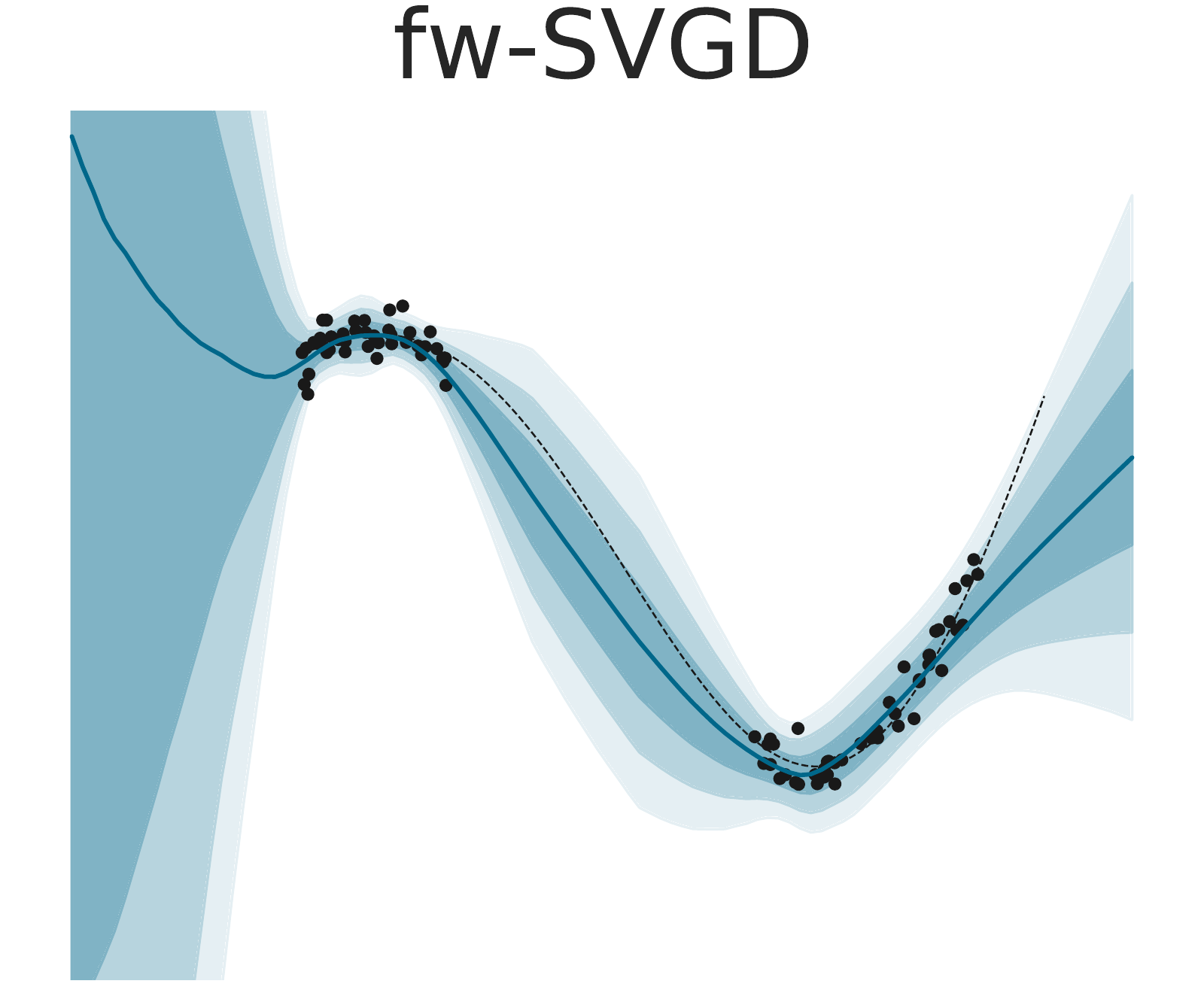}
    \end{subfigure}
    \begin{subfigure}[b]{0.18\textwidth}
    \includegraphics[width=\linewidth,trim={1cm 1cm 1cm 0cm},clip]{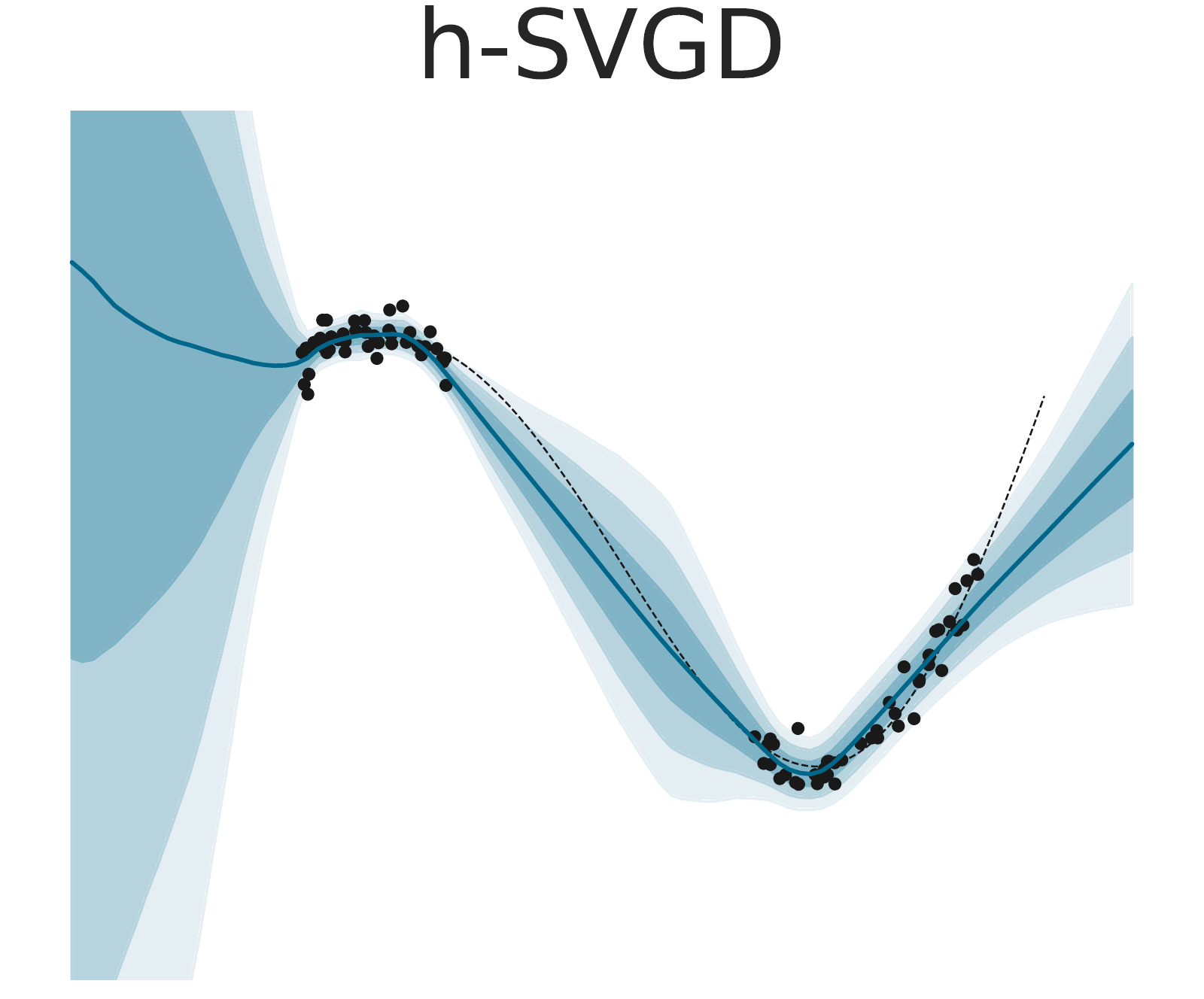}
    \end{subfigure}
    
    \hspace{3.8mm}
    \begin{subfigure}[b]{0.18\textwidth}
    \includegraphics[width=\linewidth,trim={3cm 3cm 3cm 3cm},clip]{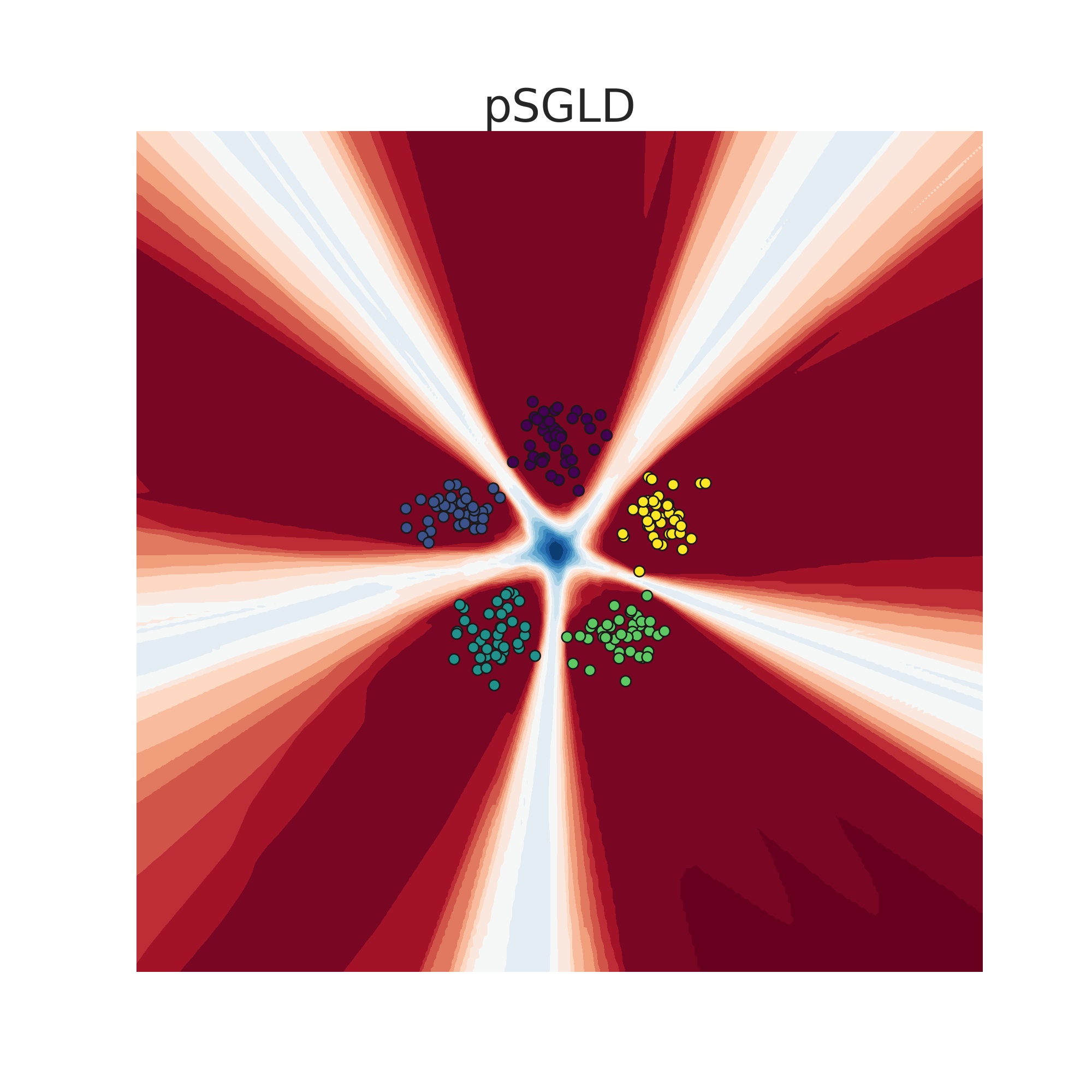}
    \end{subfigure}
    \begin{subfigure}[b]{0.18\textwidth}
    \includegraphics[width=\linewidth,trim={3cm 3cm 3cm 3cm},clip]{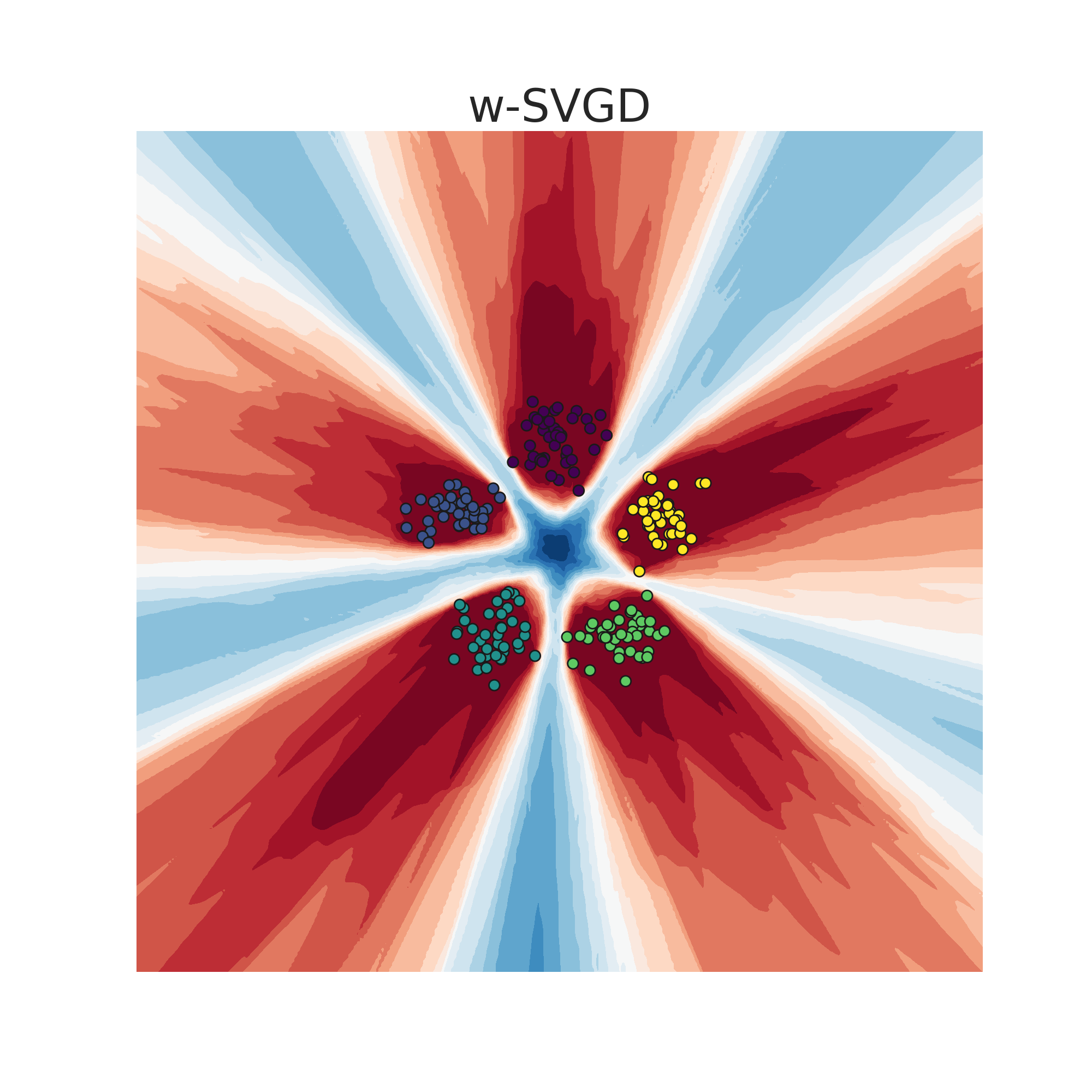}
    \end{subfigure}
    \begin{subfigure}[b]{0.18\textwidth}
    \includegraphics[width=\linewidth, trim={3cm 3cm 3cm 3cm},clip]{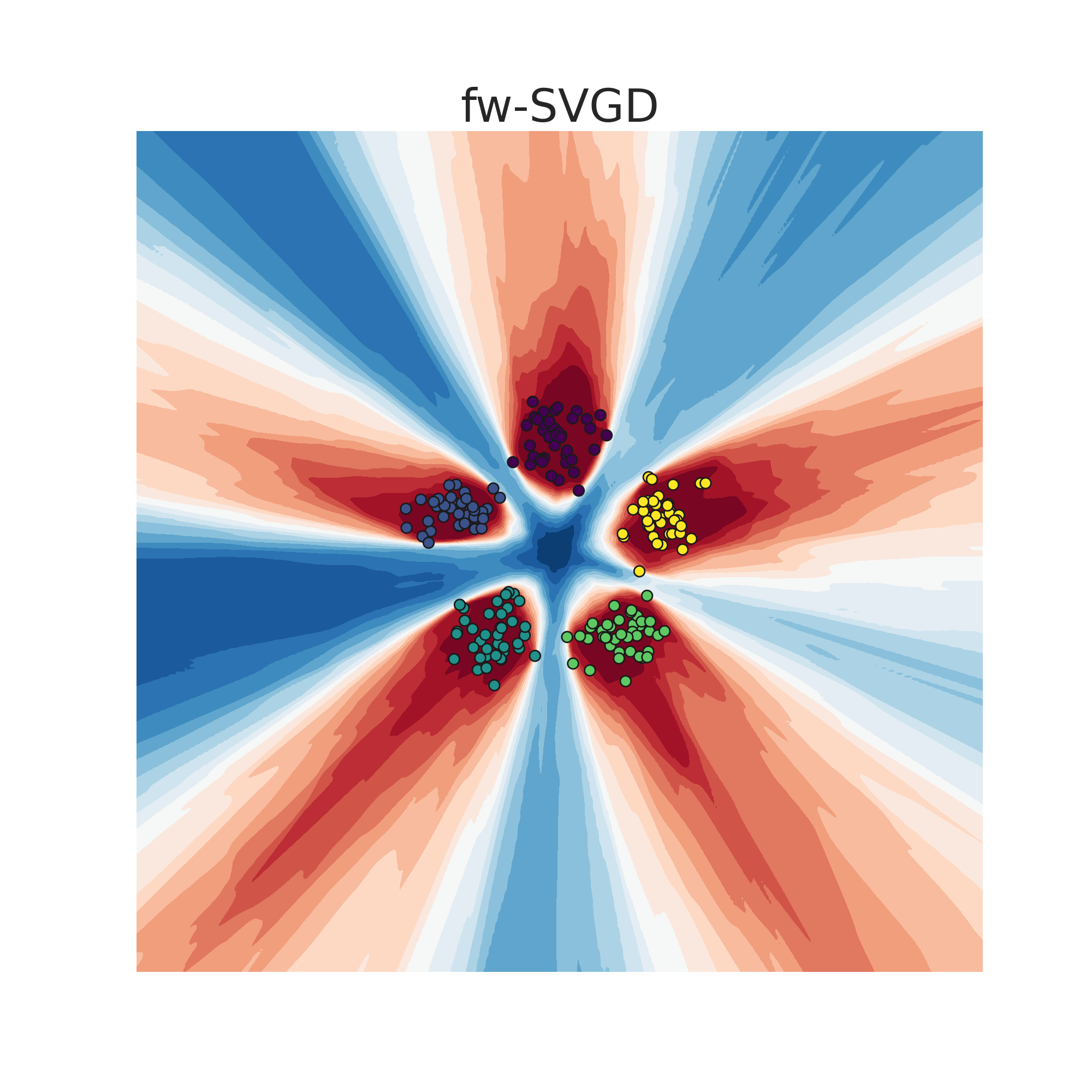}
    \end{subfigure}
    \begin{subfigure}[b]{0.18\textwidth}
    \includegraphics[width=\linewidth, trim={3cm 3cm 3cm 3cm},clip]{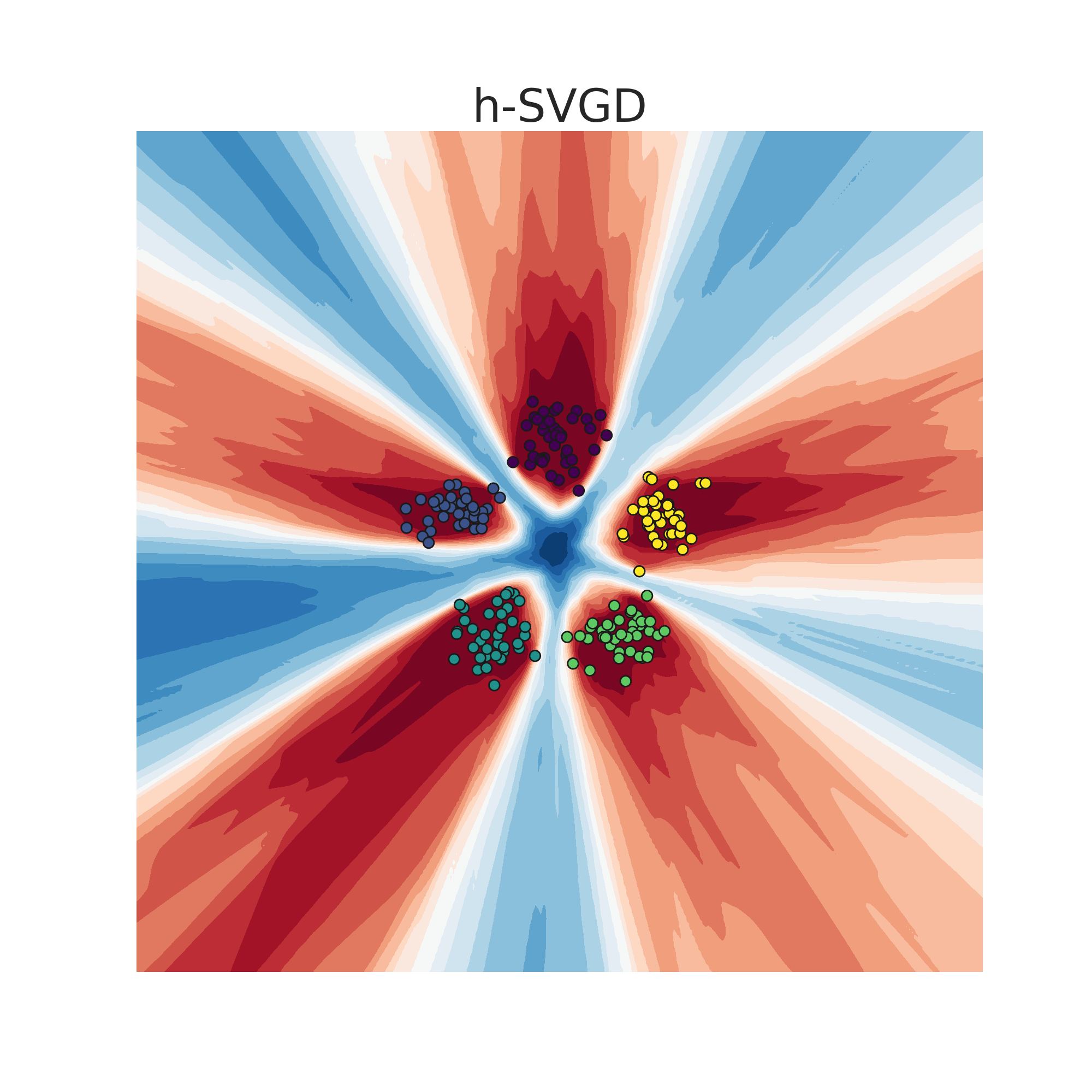}
    \end{subfigure}
    \begin{subfigure}[b]{0.032\textwidth}
    \includegraphics[width=\linewidth,trim={0.4cm 0.5cm 0cm 0.2cm},clip]{figures/cbar.pdf}
    \vspace{-5mm}
    \end{subfigure}
    
\caption{\textbf{Synthetic data, $\beta = 1$.} Predictive posteriors for the different stochastic update methods on a synthetic 1D regression (top) and 2D classification (bottom) task. Shaded areas in the regression represent the standard deviation of the predictive posteriors, for the classification we show the entropy of the predictive posteriors. We see again that the function-space methods capture the uncertainty better than the weight-space ones, but all methods generally seem to capture it better than in the case of using deterministic updates.}
\label{fig:toy_tasks_sto}
\end{figure*}

In Figure~\ref{fig:toy_tasks_sto} we report the results for the stochastic methods for the one-dimensional regression task and the two-dimensioanl classification. 

\subsection{Calibration curves}
\label{sec:calibration_curves}
In this section, following the protocol of \citep{ciosek2019conservative}, we show the calibration curves for the different methods on FashionMNIST in Figure~\ref{fig:cal_mnist} and on CIFAR10 in Figure~\ref{fig:cal_cifar}.
These curves show the average accuracy of all methods on differently sized subsets of a test dataset, which is a combination of the (FashionMNIST,CIFAR10) test set and 10,000 OOD (MNIST,SVHM) data points.
Each subset is constructed by including the $k$ points with the smallest uncertainty as predicted by the respective method. In the plots we display only the first 10,000 points sorted by confidence, to make the results better visible. This simulates a setting in which the model is deployed on data that may or may not be OOD and has to decide based on its uncertainty estimates for which points to actually make a prediction. If a model is well-calibrated, this curve should decrease monotonically when increasing the subset size.

\subsubsection{FashionMNIST}
\begin{figure}[h]
\centering
\includegraphics[width=0.49\textwidth]{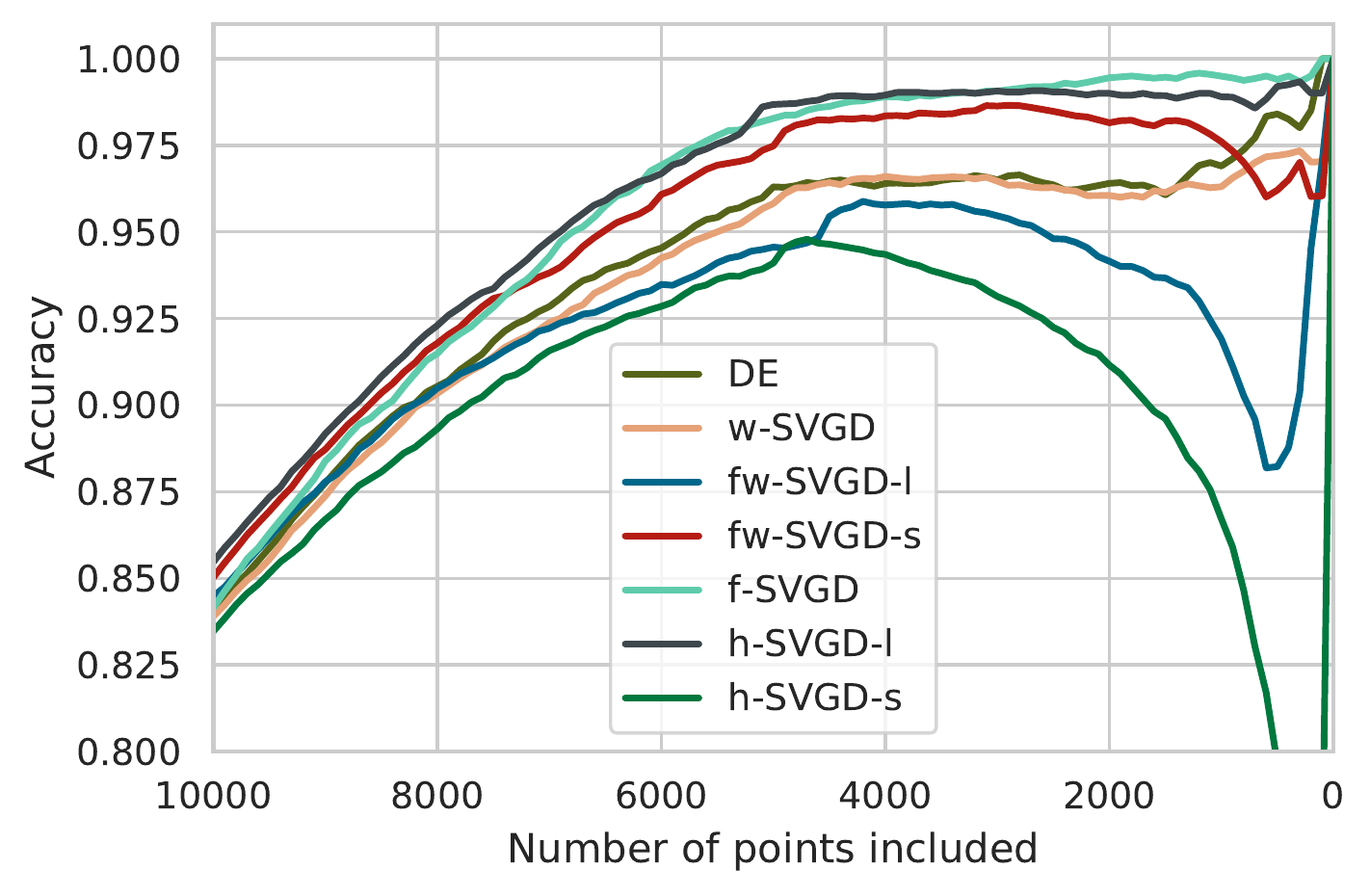}
\includegraphics[width=0.49\textwidth]{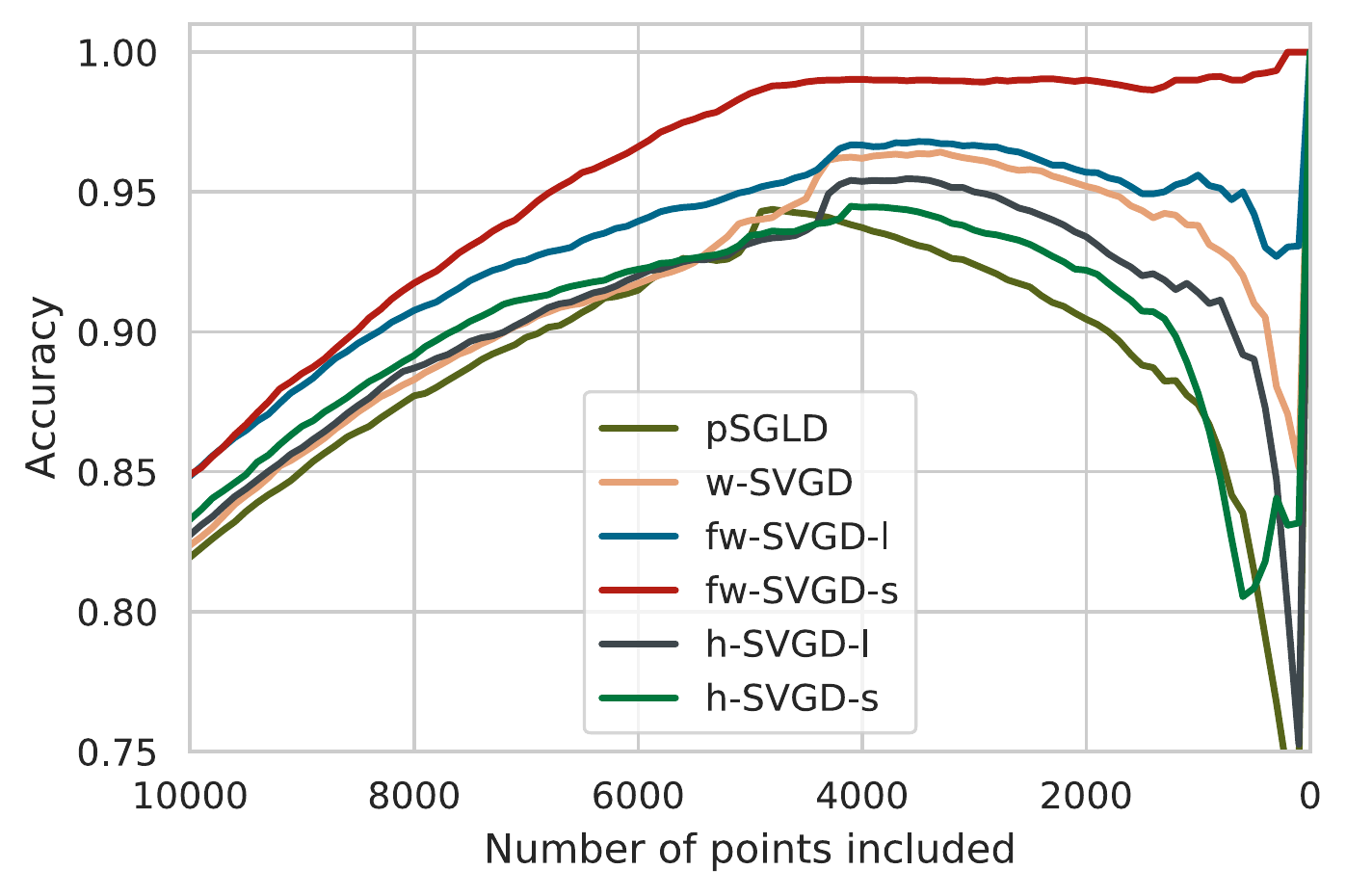}
\caption{\textbf{Calibration curves FashionMNIST:} calibration curves showing the relationship between uncertainty (horizontal axis) and accuracy (vertical axis) for the stochastic methods  $\beta = 1$ (right) and deterministic  $\beta = 0$(left).}
\label{fig:cal_mnist}
\end{figure}

We can see in Figure~\ref{fig:cal_mnist} that for f-SVGD, h-SVGD-l, w-SVGD and DE the curve is monotonically decreasing. Conversely h-SVGD-s, fw-SVGD-l and fw-SVGD-s do not display this behaviour as the accuracy is decreasing between 5000 and 0, fact that reflects low uncertainty over OOD points that are then incorrectly classified.
Furthermore, the area under this curve can be seen as a combined measure for accuracy and uncertainty calibration. For instance, when setting a target accuracy of 97~\%, the h-SVGD-l and f-SVGD in Figure~\ref{fig:cal_mnist}  could be used to classify 6,000 points which is more than for any of the other methods. 
For the stochastic case instead, fw-SVGD-s shows the best performance and appears to be the only calibrated method.

\subsubsection{CIFAR10}

\begin{figure}[h]
\centering
\includegraphics[width=0.49\textwidth]{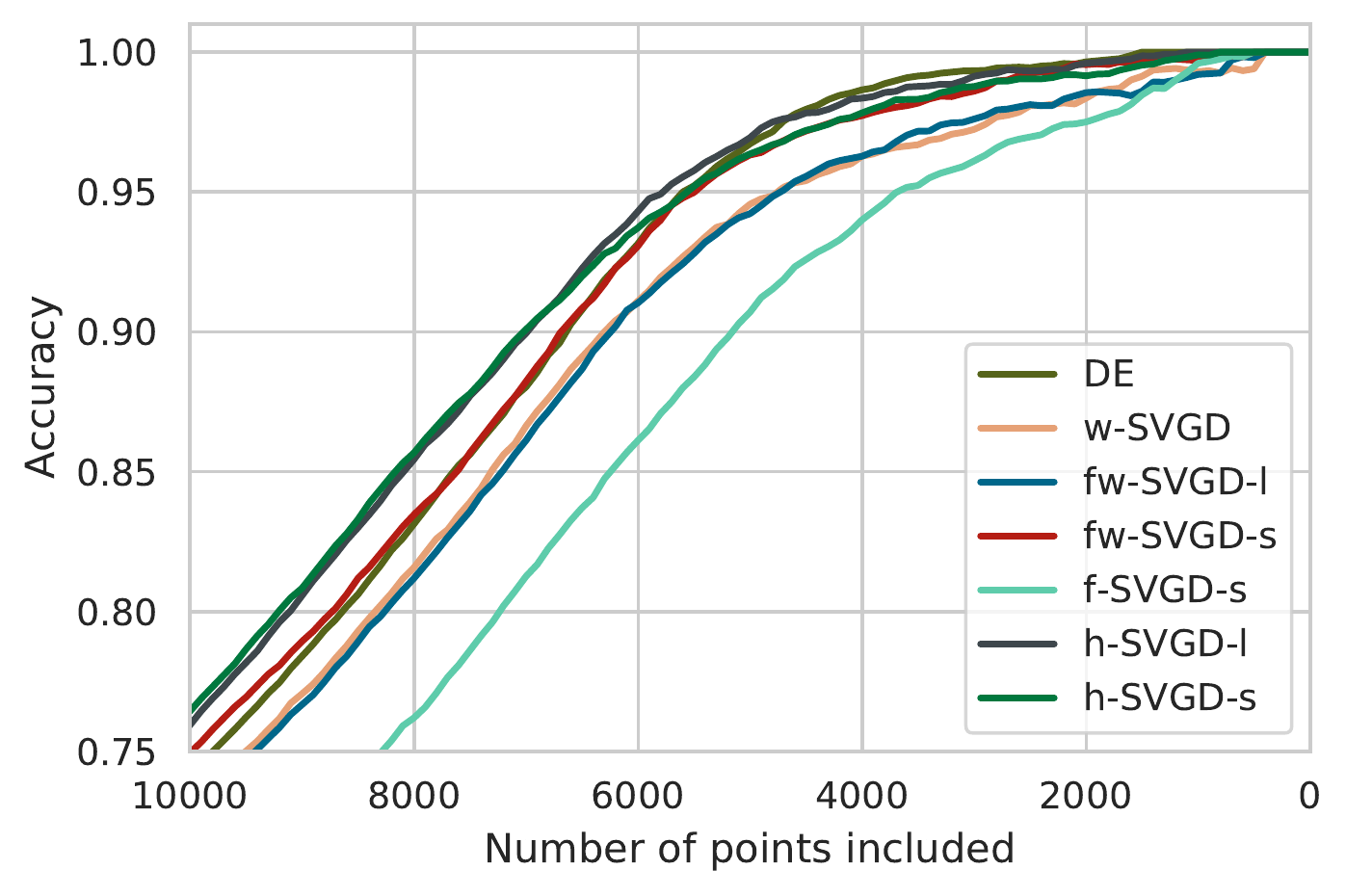}
\includegraphics[width=0.49\textwidth]{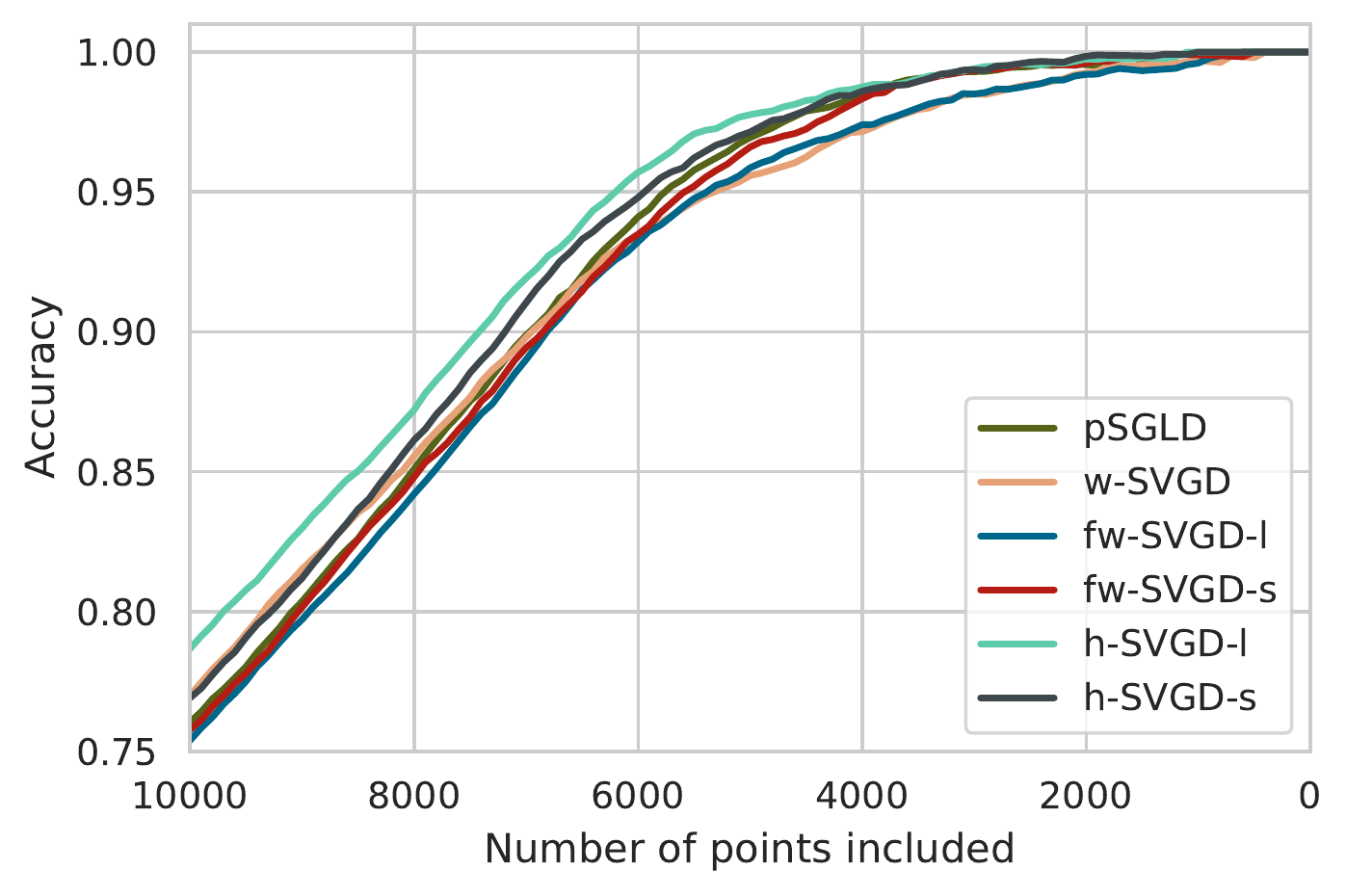}
\caption{\textbf{Calibration curves CIFAR-10:} calibration curves showing the relationship between uncertainty (horizontal axis) and
accuracy (vertical axis) for the stochastic methods  $\beta = 1$ (right) and deterministic  $\beta = 0$(left).}
\label{fig:cal_cifar}
\end{figure}

We observe in Figure~\ref{fig:cal_cifar} that on CIFAR10, the h-SVGD-l appears to be the one with the largest area under the curve and so the one achieving the best tradeoff between uncertainty and accuracy in both the deterministic and stochastic case.

\subsection{Corrupted data}
\label{sec:corruption}

\begin{figure}
\centering
    \begin{subfigure}[b]{1.1\textwidth}
        \includegraphics[width=\linewidth]{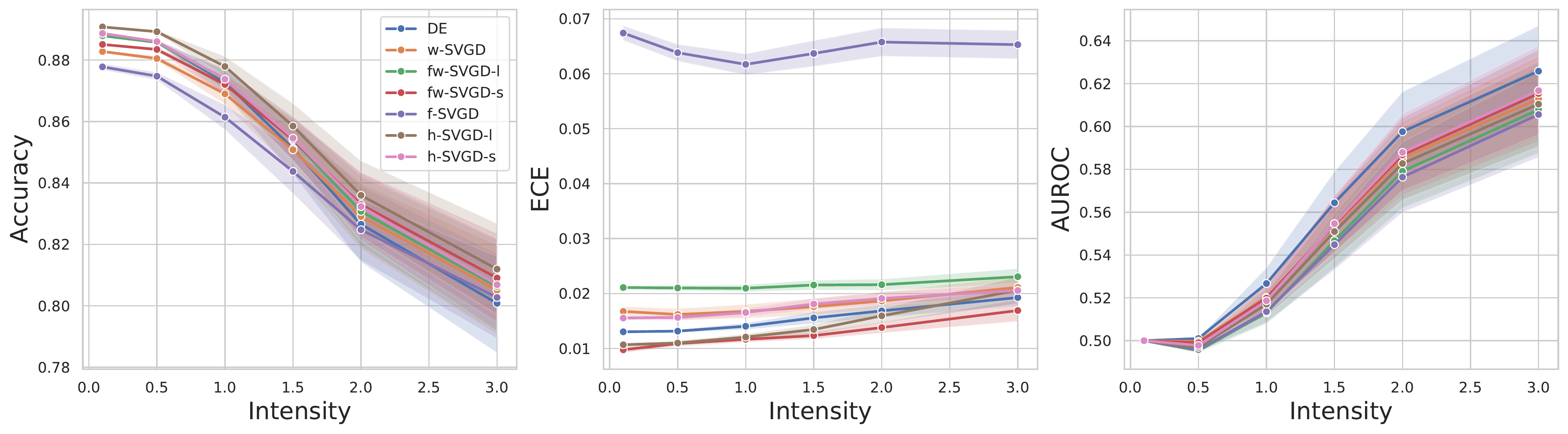}
        \caption{ \raggedright \textbf{Corruption analysis FashionMNIST $\beta = 0$.}}
        \label{fig:corr_mnist_det}
    \end{subfigure}
    \begin{subfigure}[b]{1.1\textwidth}
        \includegraphics[width=\linewidth]{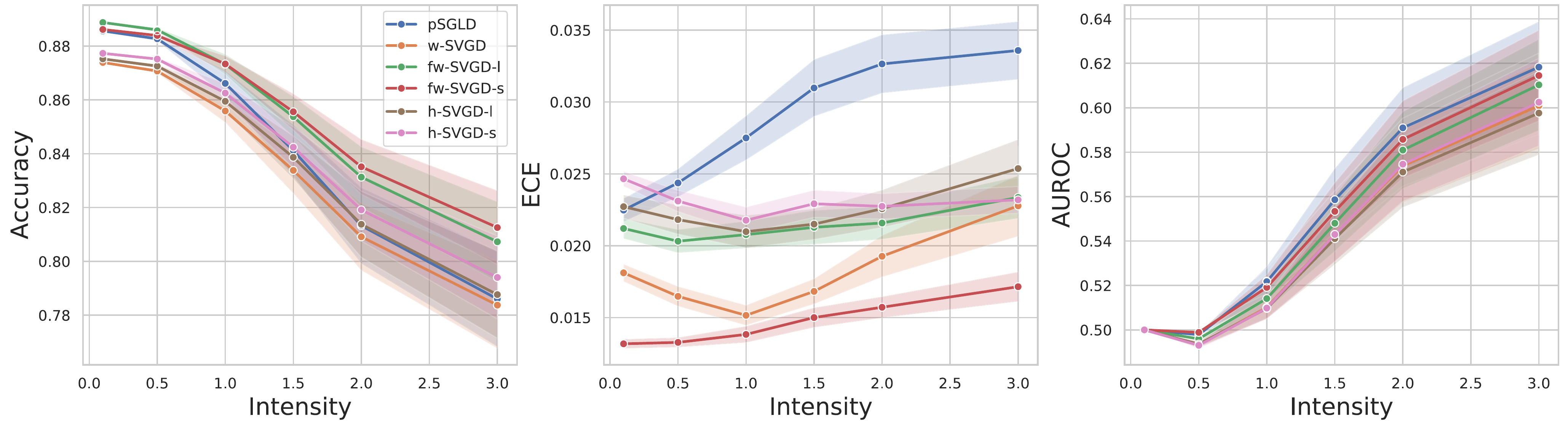}
        \caption{  \raggedright \textbf{Corruption analysis FashionMNIST $\beta = 1$.}}
        \label{fig:corr_mnist_sto}
    \end{subfigure}
    \begin{subfigure}[b]{1.1\textwidth}
        \includegraphics[width=\linewidth]{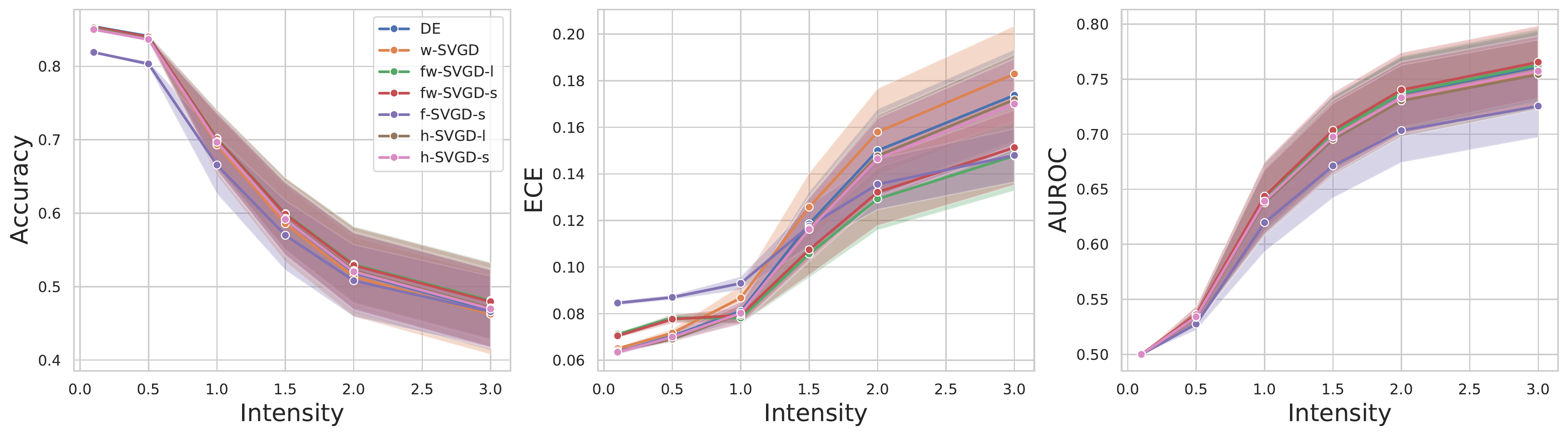}
        \caption{  \raggedright \textbf{Corruption analysis CIFAR-10 $\beta = 0$.}}
        \label{fig:corr_cifar_det}
    \end{subfigure}
    \begin{subfigure}[b]{1.1\textwidth}
        \includegraphics[width=\linewidth]{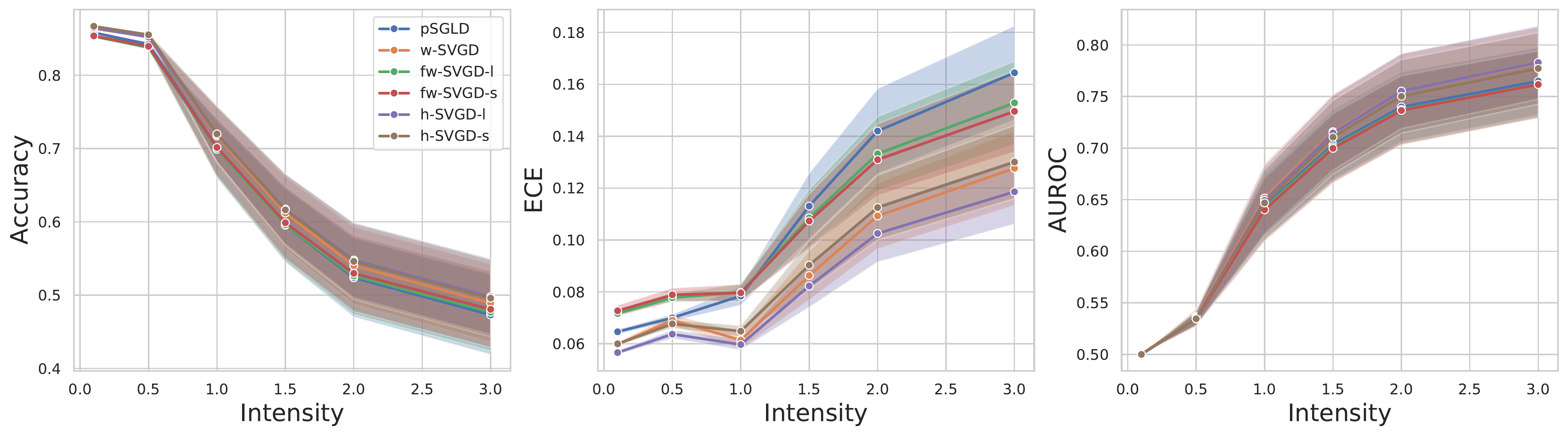}
        \caption{ \raggedright  \textbf{Corruption analysis CIFAR-10 $\beta = 1$.}}
        \label{fig:corr_cifar_sto}
    \end{subfigure}
\caption{we compare accuracy, ECE, and AUROC under corruption of the FashionMNIST and CIFAR10 dataset, given by Gaussian blur. For each method, we show the mean and standard error on the corrupted test set for different intensities of corruption.}
\end{figure}

We studied the accuracy, ECE, and AUROC score of the different methods illustrated in the main text on datapoints corrupted with different intensities. The corruption is created by blurring the images in the test set with random Gaussian blur. The intensity of the corruption is determined by the standard deviation of the Gaussian kernel used for the blurring, which is selected from $\sigma \in \{ 0.1,0.5,1,1.5,2.0,3.0\}$. The kernel size is fixed to 7. The results for FashionMNIST are reported in Figure~\ref{fig:corr_mnist_det} for the deterministic case and Figure~\ref{fig:corr_mnist_sto} for the stochastic one. The first one shows that the Deep Ensemble has a tendency of considering the corrupted data OOD; for that reason that accuracy is decreasing faster and the AUROC is the most elevated, this fact could also be interpreted as a tendency of deep ensemble to overfit. On the other side the functional methods that we introduced seem more robust in the sense that the accuracy on the corrupted data remains high, especially for the h-SVGD-l and for the same reason the AUROC is lower. Interestingly, the f-SVGD leads to less calibrated predictions when compared to all other methods. For the stochastic variants, instead, the fw-SVGD-s and fw-SVGD-l  seem to be the more robust in accuracy and the pSGLD displays characteristics very similar to the Deep Ensemble. The results for CIFAR-10 in Figure~\ref{fig:corr_cifar_det} and Figure~\ref{fig:corr_cifar_sto} show a similar trend for all methods in terms of accuracy and AUROC, the only relevant difference is in the ECE for which the fw-SVGD appears more calibrated across the corruptions for the deterministic case and the h-SVGD-l for the stochastic one. The worse performance of  f-SVGD for bigger models is confirmed also in this experiment.  


\newpage

\section{Implementation details}
\label{sec:imp_details}

In this section, we report details on our implementations in the experiments we performed. The code is partially based on \citet{oshg2019hypercl}. All the experiments were performed on an internal cluster with NVIDIA GTX 1080 Ti and took roughly 200 GPU hours.

\subsection{1D regression}
We generate the training data by sampling 45 points from $x_i \sim \text{Uniform}(1.5,2.5)$ and 45 from $x_i \sim \text{Uniform}(4.5,6.0)$. The output $y_i$ for a given $x_i$ is then modeled following $y_i = x_i\sin(x_i) + \epsilon_i$ with $\epsilon_i \sim \mathcal{N}(0,0.25)$. We use a standard Gaussian likelihood and standard normal prior $\mathcal{N}(0, \mathbb{I})$. The model is a feed-forward neural network with 2 hidden layers and 50 hidden units with ReLU activation function. We use 50 particles initialized with random samples from the prior and optimize them using Adam \citep{kingma2014adam} with 10000 gradient steps, a learning rate of 0.001 and batchsize 64. The kernel bandwidth is estimated using the median heuristic. We test the models on 100 uniformly distributed points in the interval $[0,7]$. The random seed was fixed to 42.

\subsection{2D classification} 
We generate 200 training data points sampled from a mixture of 5 Gaussians with means equidistant on a ring of radius 5 and unitary covariance. The model is a feed-forward neural network with 2 hidden layers and 50 hidden units with ReLU activation function. We use a softmax likelihood and standard normal prior $\mathcal{N}(0, \mathbb{I})$. We use 100 particles initialized with random samples from the prior and optimize them using Adam \citep{kingma2014adam} with 5000 gradient steps, a learning rate of 0.01 and batchsize 64. The kernel bandwidth is estimated using the median heuristic. We use the hyperbolic annealing in the first 2000 iterations. The random seed was fixed to 42.

\subsection{Classification on FashionMNIST} 
On this dataset, we used a feed-forward neural network with 3 hidden layers and 100 hidden units with ReLU activation function.The likelihood was softmax and the prior standard normal $\mathcal{N}(0, \mathbb{I})$. We used 50 particles initialized with random samples from the prior and optimize them using Adam \citep{kingma2014adam} for 60000 steps. For $\beta = 0$ the learning rate was 0.0025 for h-SVGD-l, h-SVGD-s, fw-SVGD-s,fw-SVGD-l,DE and 0.005 for f-SVGD, w-SVGD, we used the hyperbolic annealing \citep{dangelo2021annealed} for h-SVGD and fw-SVGD and the linear annealing for f-SVGD in the first 1000 steps. For $\beta = 1$ instead we used a learning rate of $0.0025$ for fw-SVGD-l,pSGLD, and 0.005 for fw-SVGD-s,h-SVGD-s,h-SVGD-l and w-SVGD, the hyperbolic annealing \citep{dangelo2021annealed} was applied to h-SVGD and fw-SVGD for the first 1000 steps. A batchsize of 256 was used for all experiments. The kernel bandwidth is estimated using the median heuristic for all different methods. The learning rates were searched over the following values $(1e-4,5e-4,1e-3,25e-4,5e-3)$. We tested for 60000 and 30000 total number of steps  and batchsize 256 and 128. All methods with a repulsive force were tested using hyperbolic, cyclical, linear and without annealing. All results in Table~\ref{tab:fmnist} are averaged over the following random seeds $(38,39,40,41,42)$. 

\subsection{Classification on CIFAR-10} 
On this dataset, we used a residual network (ResNet32) with ReLU activation function.  We use a softmax likelihood and standard normal prior $\mathcal{N}(0, 0.1\mathbb{I})$.  We used 20 particles initialized using He initialization \citep{he2015delving}  and optimized them using Adam \citep{kingma2014adam} for 35000 steps. For $\beta =0$ we use a lerning rate of $0.001$ for all methods apart for f-SVGD were we used $0.0001$ given that a bigger value was always leading to numerical instabilities. For $\beta = 1$ the learning rate was fixed to 0.001 for all methods, we used linear annealing only for h-SVGD-l, h-SVGD-s for the first 1000 steps. The batchsize was 128 for all experiments. The kernel bandwidth is estimated using the median heuristic for all different methods. The learning rates were searched over the following values $(1e-4,1e-3,5e-3)$ we tested for 20 and 10 particles and all methods with a repulsive force were tested using hyperbolic, cyclical, linear and without annealing. All results in Table~\ref{tab:cifar} are averaged over the following random seeds $(38,39,40,41,42)$. 

\end{document}